\documentclass[journal,final,twocolumn]{IEEEtran}
\usepackage{amssymb}
\usepackage{mathrsfs}
\usepackage[noend]{algpseudocode}
\usepackage{algorithmicx}
\usepackage{mathrsfs}
\usepackage{graphicx}
\usepackage{amsfonts}
\usepackage{amsmath}
\usepackage{epsfig}
\usepackage{multirow}
\usepackage{arydshln}
\usepackage{threeparttable}
\usepackage{float}
\usepackage{color}
\usepackage{cite}
\usepackage{array}
\usepackage{mdwmath}
\usepackage{mdwtab}
\usepackage{eqparbox}
\usepackage[tight,footnotesize]{subfigure}
\usepackage{stfloats}
\usepackage{slashbox}
\usepackage{algorithm}
\usepackage{booktabs}
\usepackage{diagbox}
\usepackage{subfigure} 
\usepackage[colorlinks=true, linkcolor=blue, citecolor=blue, urlcolor=blue]{hyperref}
\usepackage{xcolor}
\usepackage{letltxmacro}

\LetLtxMacro\originalref\ref
\LetLtxMacro\originalautoref\autoref
\LetLtxMacro\originalcite\cite

\renewcommand{\ref}[1]{%
  \hypersetup{linkcolor=blue}
  \originalref{#1}%
}
\renewcommand{\autoref}[1]{%
  \hypersetup{linkcolor=blue}
  \originalautoref{#1}%
}

\renewcommand{\cite}[1]{%
  \hypersetup{citecolor=blue}
  \originalcite{#1}%
}

\usepackage{mathrsfs}
\allowdisplaybreaks
\begin{document}

\title{ High-Precision Formation Control for Heterogeneous Multi-Robot Systems via Hierarchical Hybrid Physics-Informed Deep Reinforcement Learning}
\author{Yanzhou Li, Guangli Chen, Xiao-Meng Li, Wenjian Zhong, Yongkang Lu, Shenghuang He

\thanks{
This work is supported in part by Guangdong Basic and Applied Basic Research Foundation under Grants 2023A1515110249, 2024A1515110151, and 2024A1515140079, and in part by China Postdoctoral Science Foundation under Grant 2024M760418. (\textit{Corresponding author: Xiao-Meng Li}).

Yanzhou Li, and Wenjian Zhong are with Dongguan Key Laboratory of Intelligent Equipment and Smart Industry, School of Advanced Engineering,  Great Bay University, Dongguan 523000, China(e-mail:lyz19921207@163.com, wenjianzhong973008@163.com).

Guangli Chen is with the Chair of Applied Statistics, Technische Universit{\"a}t Dresden, Dresden, Germany (guangli.chen@mailbox.tu-dresden.de). 

Xiao-Meng Li is with the School of Mechanical and Electrical Engineering, Guangzhou University, Guangzhou 510006, China (e-mail: lixiaomeng66@163.com).


Yongkang Lu is with Dongguan Key Laboratory of Intelligent Equipment and Smart Industry, School of Advanced Engineering,  Great Bay University, Dongguan 523000, China, and is also with University of Science and Technology of China, Hefei, 230026 (e-mail:lykang123@gbu.edu.cn).

Shenghuang He is with the College of Computer Science, Dongguan University of Technology, Dongguan 523808, China (e-mail:shhe6688@163.com).
}
}
\maketitle

\begin{abstract} 
Existing  classical control methods commonly require precise models and  struggle to cope with model uncertainties and external disturbances, while end-to-end reinforcement learning (RL) approaches suffer from low sample efficiency and poor convergence.  To overcome these challenges, 
this paper proposes a hierarchical hybrid physics-informed deep reinforcement learning (HHy-PIDRL) framework, aiming to realize high-precision, highly responsive formation control for heterogeneous multi-robot systems (HMRSs).  The proposed framework contains two layers. Specifically, 
first, the upper layer  designs an autonomous navigation policy network for Ackermann-steering leader based on the Soft Actor-Critic (SAC) deep reinforcement learning (DRL) algorithm. 
Second, the lower module integrates a high-fidelity physical feed-forward controller, a classical proportional-derivative (PD) controller, and an adaptive DRL residual controller to propose an effective hybrid model and DRL (HM-DRL)-based formation control policy network. 
Third, a unique hierarchical reward function is designed for training Omnidirectional followers, which effectively guides agents toward a refined, stable control policy. 
Experimental results demonstrate that, the success rate of both the  upper-layer autonomous navigation policy network and the  HM-DRL based formation control policy networks reach $100\%$. Meanwhile, ablation experiments are conducted to verify the validity and credibility of the proposed method. 
\end{abstract}




\IEEEpeerreviewmaketitle

\section{Introduction}
Multi-robot systems (MRSs) have become a cornerstone of modern robotics, enabling complex, coordinated tasks across domains such as  environmental monitoring \cite{zhu2024multi} search and rescue \cite{wang2023development} and intelligent transportation \cite{chen2025reinforced}. Among the key technologies for MRSs, formation control stands out as a fundamental problem \cite{sakurama2024formation,li2025deform }. In some scenarios, compared with homogeneous MRSs, heterogeneous multi-robot systems (HMRSs) possess significant advantages by leveraging the unique strengths of different agents to achieve complex objectives with greater efficiency, robustness, and cost-effectiveness \cite{zhang2023heterogeneous,xiao2025leader}. Thus, research on HMRSs has attracted significant attention and investigation, with formation control emerging as one of the key research areas. However, considering that heterogeneous robots possess different kinematic constraints, research on the formation control of HMRSs faces significant challenge. 
The challenge is  amplified in scenarios involving a leader with nonholonomic constraints, such as an Ackerman-steering vehicle, followed by more agile agents, such as Omnidirectional platforms. The leader's nonlinear and coupled dynamics, especially during high-speed and high-curvature maneuvers, introduce complex centripetal and Euler forces that propagate through the formation. These dynamics create a highly non-stationary control problem for the followers, rendering traditional linear control methods insufficient for high-fidelity tracking.

Until now, several approaches, such as  proportional-derivative (PD) control, model predictive control (MPC) and  end-to-end deep reinforcement learning (DRL)-based control have been developed to achieve the formation control. 
For example,  Wang \textit{et al}. \cite{farivarnejad2020decentralized} addressed the lag consensus problem of multiagent systems based on PD and proportional-integral (PI) control. Nayak \textit{et al}. \cite{nayak2023heuristic} utilized PD controller to generate safety trajectories.  Luis \textit{et al}. \cite{luis2020online} developed a MPC algorithm to generate trajectories in real time for multiple robots. Jiang \textit{et al}. \cite{jiang2023incorporating} developed a distributed model predictive controller to achieve  coordinated control of multiple robots.  Unfortunately, these PD-based and MPC-based  control approaches often require precise system models and struggle to adapt to unmodeled dynamics or external disturbances. Conversely, end-to-end deep reinforcement learning (DRL) approaches have demonstrated remarkable success in learning complex control policies from data.  Zhong \textit{et al}. \cite{zhong2025distributed} proposed a distributed DRL approach for the formation control of Ackermann MRSs.  Li \textit{et al}. \cite{li2024distributed} developed a bi-objective DRL framework for the formation control of MRSs, and comparative experiments were carried out to verify the effectiveness of the designed framework. Other  DRL-based formation control approaches for MRSs can refer to the references \cite{xing2024multi,dawood2025safe}. 
However, these ``black-box'' methods are notoriously sample-inefficient and often fail to leverage the well-understood physics of the system. This can lead to unstable learning processes and policies that are not robust, particularly when precise formation tracking is required. 

To address these limitations, this paper develops an effective hierarchical hybrid physics-informed deep reinforcement learning (HHy-PIDRL) framework for the formation control of HMRSs. The proposed framework contains two layers, and synergistically decomposes the control problem, integrating the strengths of classical control, physics-based models, and data-driven learning. The upper layer is the single Ackermann-steering robot navigation module and the lower layer is the multi Omnidirectional robot formation tracking module. Both layers are trained by Soft Actor-Critic (SAC)-based DRL algorithm. 
The main contributions of this paper are given as follows:

$\bullet$ $\mathbf{HHy}$-$\mathbf{PIDRL~ Framework}$: The complex nonlinear formation control problem of HMRSs is decomposed into two layers, which demonstrates considerable flexibility and practicality. For lower layer, a physics-based feedforward and PD controllers provide a robust, stable, and interpretable control base. An adaptive RL residual controller is designed to learn a simpler, lower-dimensional residual function. This framework dramatically reduces the agent's exploration space, significantly improving learning efficiency and final control performance. 

$\bullet$  $\mathbf{Hierarchical~Reward~Shaping}$: The reward function design embodies a hierarchical guidance from ``approaching'' to ``holding''. The introduction of the high-precision holding bonus specifically addresses the common vanishing gradient problem when the error is small, giving the agents the incentive to pursue maximum stability and precision. 
    
$\bullet$ $\mathbf{Information}$-$\mathbf{Enhanced~State~Representation}$: The agent's state space innovatively  incorporates the global formation average error and its own historical error sequence. The former provides each distributed agent with a global perspective, fostering more coordinated group behavior. The latter allows the agent to perceive error trends, enhancing the predictability of its decision-making. 

\section{related work}\label{sec:related_work}
This paper focuses on two core modules: single-robot autonomous navigation and multi-robot formation control. Below is a review of the relevant literature for these two modules. 

\subsection{Single Robot Autonomous Navigation}
Single-robot navigation is a fundamental challenge in mobile robotics, aimed at directing a robot to traverse safely and effectively from an origin to a destination  \cite{xu2024research, xie2025robot,chen2023self,pore2023autonomous,sawant2023hybrid}. Based on the core ideas of technological evolution and dependence, relevant research work can be divided into the following two categories: classic hierarchical planning methods and learning-based methods. 

$\bullet$ $\mathbf{Classic~Hierarchical~Planning~Methods}$: The hierarchical perception-planning-control architecture is widely adopted in robotic navigation systems. The core idea is to decompose the navigation task into four key subproblems: localization, mapping, global path planning, and local path planning. For example, global path planning typically employs classical algorithms such as $A^{*}$ and its variants \cite{xu2024research, xie2025robot}, $RRT$ and improved version \cite{tu2024improved,yin2023efficient}, and Dijkstra algorithms \cite{liu2023cdt}, while local planning often relies on the dynamic window approach (DWA) \cite{gu2025research,kumar2024simultaneous}. These conventional methods exhibit a well-defined structure, strong interpretability, and considerable technological maturity, making them the predominant approaches in current industrial applications. However, their performance is heavily dependent on accurate environmental maps and precise robot models. 

$\bullet$ $\mathbf{Learning}$-$\mathbf{Based~Methods}$: By interacting with the environment through trial and error, the robot independently learns the mapping strategy from sensor inputs to control commands based on the  received reward signals. To effectively enhance the safety performance of robotic systems, various learning approaches have been employed, such as  supervised learning \cite{wang2024imperative}, self-supervised learning \cite{kahn2021badgr}, and RL\cite{yin2024autonomous}. Nevertheless, supervised and self-supervised methods typically require large-scale datasets to train a well-performing network policy.  In contrast, RL enables the robot to continuously improve its policy through repeated environmental interaction, thereby learning an optimized strategy network. 

This paper utilizes  SAC-based DRL algorithm to train Ackerman-steering  leader. Compared with existing results, the single robot autonomous navigation module incorporates realistic nonholonomic kinematic and dynamic models instead of simplified particle models, thus yielding greater practical applicability.

\subsection{Multi-Robot Formation}
Up to now, many methods were developed to achieve the formation control of multiple robots, and can be summarized into the following categories: 

$\bullet$ $\mathbf{Consensus}$-$\mathbf{Based~Method}$: Consensus-based formation control method is one of the most important and mainstream methods in the field of distributed control. By designing a distributed formation control protocol, the entire MRSs can finally converge to the globally expected formation state \cite{dai2023distributed}.  Although these consensus-based formation control methods theoretically solve the robot formation control problem, they require the satisfaction strict linear matrix inequality (LMI) condition. 

$\bullet$ $\mathbf{Virtual~Structure~Method}$: By treating the entire robot formation as a single rigid virtual structure, with each robot assigned a fixed desired position relative to that structure, the control objective is to drive all robots to accurately track the motion trajectories of their corresponding points on the virtual structure \cite{ouyang2023formation}. While this approach provides clear geometric relationships and stable formation keeping, it imposes stringent requirements on the trajectory tracking performance of individual robots and offers relatively limited flexibility in adapting to dynamic environmental constraints. 

$\bullet$ $\mathbf{Artificial~Potential~Field (APF)~Method}$: The core principle of the artificial potential field (APF) method for formation control is to integrate all predefined desired behaviors and constraints, such as formation keeping, obstacle avoidance, and inter-robot coordination into a unified artificial potential field. The motion of each robot is then governed by the resultant force derived from the negative gradient of this composite potential field  \cite{zhao2025formation,pan2021improved}. However,  a well-known limitation of this approach is its susceptibility to local minima, which can cause the formation to become trapped in suboptimal configurations. 

$\bullet$ $\mathbf{Behavior}$-$\mathbf{Based~Approach}$: For each robot,  a set of fundamental behaviors is designed. The final control command for the robot is synthesized as a weighted combination of the outputs from these individual behaviors \cite{wang2023bearing,lee2018decentralized}.  While this method offers modularity and intuitive design, it presents several challenges: rigorous mathematical stability analysis is often difficult to perform, the accuracy of the formation is not easily guaranteed, and the weighting of behaviors typically relies heavily on empirical tuning.

DRL methods has been extensively employed to address multi-robot formation control challenges \cite{zhu2024multi,zhang2023heterogeneous},  primarily due to its model-free nature and strong adaptability in complex, unstructured environments. However, they also face the problems of high training cost and low sample efficiency. To address these limitations, this paper aims to combine DRL and classic control methods to achieve  high-precision formation control of HMRSs with a  Ackermann-steering leader and Omnidirectional followers. 

\vspace{-0.5em}
\section{kinematic models}\label{sec:kinematic_models}
\begin{figure}[!htp]
\centering
\includegraphics[scale=0.25]{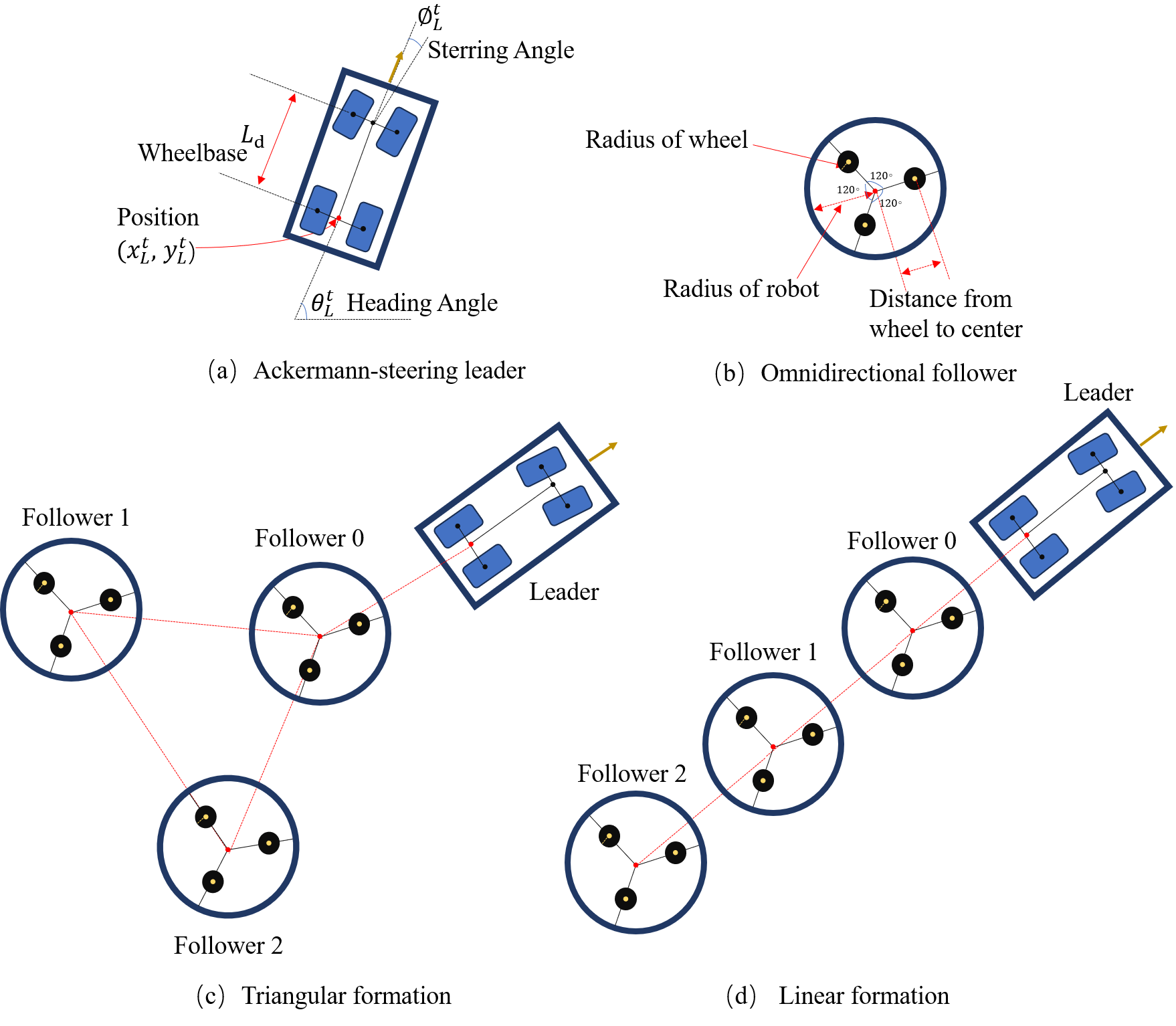}
\caption{Kinematic models of the leader and followers, and the formation patterns. (a) and (b) show the kinematic models of the leader and followers, respectively. (c) and (d)  give the formation patterns that can be achieved in this paper. }
\label{fig1}
\end{figure}

This paper investigates the formation control of HMRSs with an Ackermann-steering leader and Omnidirectional followers. The kinematic models of the leader and followers are shown in Fig. \ref{fig1}(a)  and Fig. \ref{fig1}(b), repsectively. Fig. \ref{fig1}(c)  and Fig. \ref{fig1}(d) show the triangular
formation and line formation patterns, respectively. The detailed introduction of the Ackermann-steering leader and Omnidirectional followers is given as follows.

\subsection{Kinematic Model of Leader}
Suppose the global world coordinate system is denotes as $O_{W}$: $[X_{W}, Y_{W}]$, and the local coordinate system of the leader is denotes as $O_{L}$: $[X_{L}, Y_{L}]$. The state of the leader in world coordinate system $O_{W}$ is $P_{L}^{t}=[x_{L}^{t},y_{L}^{t},\theta_{L}^{t}]^{T}$ at time $t$, where $(x_{L}^{t},y_{L}^{t})$ represents the position in 2-D plane, and $\theta_{L}^{t}\in [-\pi, \pi]$ is the heading angle. Then, the kinematic model of leader is written as 
\begin{equation}\label{E1}
{\setlength\abovedisplayskip{1pt}
\setlength\belowdisplayskip{1pt}
\begin{bmatrix}
\dot{x}_{L}^{t}\\
\dot{y}_{L}^{t}\\
\dot{\theta}_{L}^{t}
\end{bmatrix}=
\begin{bmatrix}
v_{L}^{t}cos(\mathrm{\theta}_{L}^{t})\\
v_{L}^{t}sin(\mathrm{\theta}_{L}^{t})\\
v_{L}^{t}tan(\phi^{t}_{L})/L_{d}
\end{bmatrix}}
\end{equation}
where $v_{L}^{t}$ is velocity, $\phi^{t}_{L}$ denotes the steering angle of the front wheel, $L_{d}$ is the wheelbase, i.e., the distance between the front axle and rear axle. Note that steering angle $\phi^{t}_{L}$ is bounded, which satisfies  $\phi^{t}_{L}\in[-\phi_{\mathrm{max}},\phi_{\mathrm{max}}]$. According to the physical constraints, the linear velocity is bounded, which is represented as  $v_{L}^{t}\in (0, v_{L_{\mathrm{max}}}]$, $v_{L_{\mathrm{max}}}$ is the maximum  velocity.

\subsection{Kinematic Model of Followers}
The position and speed of the Omnidirectional followers in global coordinate system $O_{W}$ are represented as $P_{F,t}^{i}=[x_{F,t}^{i},y_{F,t}^{i}]^{T}$ and $V_{F}^{i}=[v_{Fx,t}^{i},v_{Fy,t}^{i}]^{T}$, $i=1,2,\ldots,N$, respectively. $v_{Fx,t}^{i}$ and $v_{Fx,t}^{i}$ represent the velocity along the X-axis direction and Y-axis direction, respectively. Based on the movement characteristics of the omnidirectional wheel model, the kinematic model of followers can be discretized as a simple Euler integration that is described as 
\begin{equation}\label{E2}
P_{F,t+\Delta t}^{i}=P_{F,t}^{i}+V_{F,t}^{i}\Delta t
\end{equation}

Specifically, for the X-axis and Y-axis, (\ref{E2}) can be 
\begin{align}\label{E3}
x_{F,t+\Delta t}^{i}=x_{F,t}^{i}+v_{Fx,t}^{i}\Delta t\\
y_{F,t+\Delta t}^{i}=y_{F,t}^{i}+v_{Fy,t}^{i}\Delta t
\end{align}
where $\Delta t$ is the sampling interval. 
\vspace{-0.5em}
\section{PROBLEM FORMULATION}\label{sec:problem_formulation}
Multi-robot formation control is one of the important research directions in the field of multi-robot collaborative control. Traditional formation control methods and end-to-end reinforcement learning methods each have their own disadvantages. This paper aims to propose a HHy-PIDRL framework to achieve  the high-precision formation control of HMRS. The control objective is to force the Omnidirectional followers to  track the Ackermann-steering leader while maintain certain preset formation, the kinematic model of the leader and followers are shown in Sec. \ref{sec:kinematic_models}. The initial position $P_{L}^{0}=[x_{L}^{0},y_{L}^{0},\theta_{L}^{0}]^{T}$ and target position $P_{L}^{T}=[x_{L}^{T},y_{L}^{T},\theta_{L}^{T}]^{T}$ of the leader is random, and 
SAC-based RL is implemented to train the leader to reach the target position. The initial positions of all followers surround the leader in a certain shape. 
Then, HM-DRL based formation control law is proposed to control the Omnidirectional followers to tracking the leader from the initial $P_{L}^{0}$ to the target position $P_{L}^{T}$.  Without loss of generality, it is assumed that each robot can locate itself using sensors without prior knowledge of the environment, and that there is no need for communication between robots. 

\begin{figure*}
\centering
\includegraphics[scale=0.35]{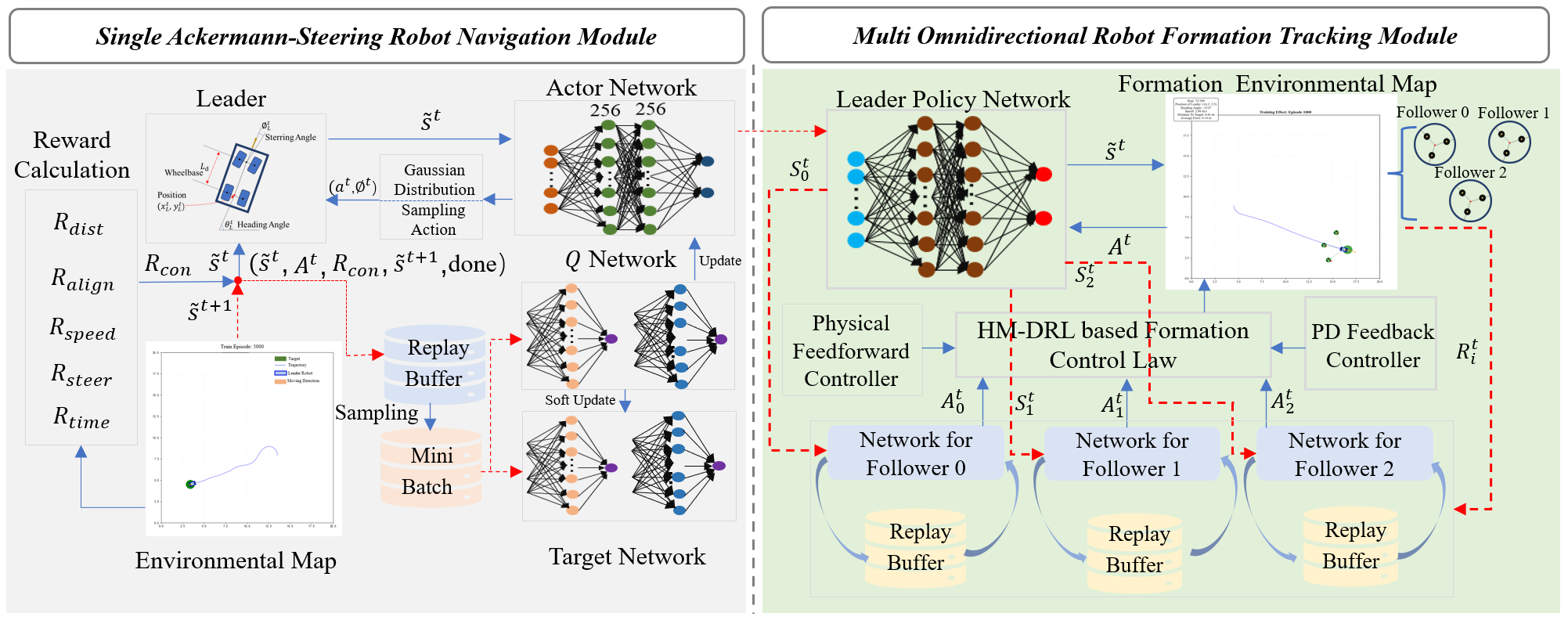}
\caption{The proposed HHy-PIDRL framework.The proposed HHy-PIDRL framework consists of two core modules, as illustrated in the accompanying diagram: the autonomous navigation module for a single Ackermann‑steering robot (gray area) and the formation‑tracking module for multiple Omnidirectional robots (green area). 
The first module is designed to train a policy network that enables the leader robot to navigate independently to designated targets. Building upon this trained leader policy, the second module introduces a HM‑DRL‑based formation control law to achieve high‑precision formation control for the  HMRSs.}
\label{fig2}
\end{figure*}

\section{PROPOSED METHODOLOGY}\label{sec:methodology}
This section aims to propose a novel HHy-PIDRL framework (shown in Fig. \ref{fig2}). The framework contains two modules:  1) \textit{Single Ackermann-steering robot navigation module}:  This module aims to train an autonomous navigation policy network using SAC-based RL algorithm,  
and 2)  \textit{Multi Omnidirectional robot formation tracking  module}: This module aims to train a HM-DRL based formation control policy network to achieve high-precision formation tracking. Subsequently, a comprehensive elaboration on these two modules is presented. 
\vspace{-1.0em}
\subsection{Single Ackermann-Steering Robot Navigation Module}
In this subsection, SAC-based RL algorithm is designed to train the Ackermann-steering leader robot to reach any target positions from random initial positions. The leader robot control process can be considered as a Markov decision process (MDP) that is consist of five-tuple $(S,A,P,R,\gamma)$. 
$S\in \mathbf{R}^{m}$ denotes the set of all possible  environmental states, $A\in \mathbf{R}^{n}$ is the set of all actions that the robot can perform in each state, policy $\pi(s^{t})$ select an action $a^{t}$ at state $s^{t}$, which is described as $a^{t}\sim \pi(s^{t})$.  $P(s^{'}|s,a): S\times A\times A\longrightarrow [0,1]$ is the state transition probability representing the probability of transferring from state $s$ to state $s^{'}$ after performing action $a$ under state $s$. $R(s,a)$ represents the reward obtained after performing action $a$ under state $s$. $\gamma$ represents the discount rate for cumulative returns, with a value range of $[0,1]$. $\gamma$  is utilized to balance the importance of current rewards and future rewards. The closer $\gamma$ is to $1$, the more the robot focuses on long-term rewards; the closer $\gamma$ is to $0$, the more the robot focuses on short-term rewards. Subsequently, the state space, action space, and reward function dedicated to the training of the Ackermann-steering leader robot are formally defined. 

\subsubsection{State Space}The state space of the Ackermann-steering leader robot is a eight-dimensional continuous state space, and is defined as 
$S^{t}_{L}=[x^{t}_{L}, y^{t}_{L}, \theta_{L}^{t}, v_{L}^{t}, d_{x}^{t}, d_{y}^{t}, D^{t}, \psi^{t}]$. $x^{t}_{L}$ and $y^{t}_{L}$ denote the current position in global coordinate system $O_{W}$, $\theta_{L}^{t}$ represents current  heading angle, $v_{L}^{t}$ is current velocity, 
$d_{x}^{t}=x_{L}^{T}-x_{L}^{t}$ and $d_{y}^{t}=y_{L}^{T}-y_{L}^{t}$ represent the relative distance of target position relative to current position,  $D^{t}=\sqrt{(d_{x}^{t})^{2}+(d_{y}^{t})^{2}}$ denotes the Euclidean distance of current position relative to the target position, 
$\psi^{t}=(\mathrm{atan}2(d_{y}^{t},d_{x}^{t})-\theta_{L}^{t}+\pi)(\mathrm{mod}(2\pi))-\pi$ denotes the relative angle to the target position, where $\mathrm{mod}(\cdot)$ denotes a mathematical operation. To improve training efficiency, the state space variables were normalized and denoted as $\tilde{S}^{t}=[x^{t}_{L}/(2O_{\mathrm{wid}}-1), y^{t}_{L}/(2O_{\mathrm{hei}}-1), \theta_{L}^{t}/\pi, v_{L}^{t}/v_{L_{\mathrm{max}}}, d_{x}^{t}/O_{\mathrm{wid}}, d_{y}^{t}/O_{\mathrm{hei}}$,  $D^{t}/\sqrt{O_{\mathrm{wid}}^{2}+O_{\mathrm{hei}}^{2}}, \psi^{t}/\pi]$, where $O_{\mathrm{wid}}$ and  $O_{\mathrm{hei}}$ represent the width and height of the environment. 

\subsubsection{Action Space} The two-dimensional continuous action space is defined as $A^{t}_{L}=[a^{t}_{L}, \phi^{t}_{L}]$, where $a^{t}_{L}$ is the normalized acceleration command and $\phi^{t}_{L}$ is the normalized steering command. Note that the normalized steering command is  constrained
the maximum rate of change of the steering angle, this makes the steering action smoother rather than being completed instantaneously. 

\subsubsection{Reward Functions} The designed reward functions (\ref{E5})--(\ref{E9}) consist of the continuous reward at each step and the sparse reward at the end of the episode. The designed per-step continuous reward function contain five aspect:
\begin{align}\label{E5}
R_{\mathrm{con}}=R_{\mathrm{dist}}+R_{\mathrm{align}}+R_{\mathrm{speed}}+R_{\mathrm{steer}}+R_{\mathrm{time}}
\end{align}

$\bullet$  Distance reward $R_{\mathrm{dist}}$: Distance reward  $R_{\mathrm{dist}}$ is an exponential decay reward, and  designed to derive the leader to quickly approach the target. The closer it gets to the target position, the higher the reward it  receive. Distance reward  $R_{\mathrm{dist}}$ is designed as 
\begin{align}\label{E6}
R_{\mathrm{dist}}=10\mathrm{exp}(-0.2D^{t})
\end{align}
where $D^{t}$ denotes the distance of current position relative to the target position. 

$\bullet$ Directional alignment reward $R_{\mathrm{\mathrm{align}}}$: Reward $R_{\mathrm{\mathrm{align}}}$ is designed to encourage the robot to continuously approach the target. $R_{\mathrm{\mathrm{align}}}$ is shown as follows:
\begin{align}\label{eq:align_reward}
R_{\mathrm{align}}=8\mathrm{max}(0,\frac{\pi-|\Delta\theta_{L}^{t}|}{\pi})
\end{align}
where $|\Delta\theta_{L}^{t}|$ is the absolute angular error between the robot’s heading and the target direction. 

$\bullet$ Speed reward $R_{\text{speed}}$: $R_{\text{speed}}$ is designed to encourage the agent to maintain an ideal speed range and penalise being too slow or stopping. $R_{\text{speed}}$ is denoted as 
\begin{equation}\label{eq:speed_reward}
R_{\text{speed}}=\left\{
	\begin{aligned}
	0.5v_{L}^{t}, \quad &v_{L}^{t}\in [0.3v_{L_{\mathrm{max}}}, 0.8v_{L_{\mathrm{max}}}]\\
	-5, \quad &v_{L}^{t}\in [0.1, 0.1v_{L_{\mathrm{max}}}]\\
	-50, \quad &v_{L}^{t}<0.1\\
	\end{aligned}
	\right
.
\end{equation}

$\bullet$ Smooth steering reward $R_{\text{steer}}$: $R_{\text{steer}}$ is designed to punish harsh steering actions and encourage smooth movement, and is designed as
\begin{align}\label{E9}
R_{\text{steer}}=-1.0|\Delta \psi^{t}| 
\end{align}
where is the absolute value of the error between the current steering angle and the previous steering angle.

$\bullet$ Time penalty $R_{\text{time}}$: $R_{\text{time}}$ is designed as $R_{\text{time}}=-0.2$. Each time it moves an extra step, it receives a small negative reward to encourage the agent to complete the task quickly.

Given that the objective is to navigate to random target points, it is necessary to assign a reward to the robot upon reaching the target point. This reward is defined as $R_{\text{reach}}=1000$ if $D^{t}<r_{\mathrm{target}}$, where  $r_{\mathrm{target}}$ is radius of the target point. It should be noted that during the training phase, the episode is immediately terminated if the leader collides with the boundary or achieves the target position in any given episode.

(4) $\mathbf{Network~Training}$: An advanced SAC-based DRL algorithm is implemented to train the leader. SAC mainly contain 
a policy network that outputs a Gaussian distribution, two soft Q-networks for clipped double Q-learning to mitigate overestimation, and both a value network and a corresponding target value network to stabilize training. 
More detailed introduction of SAC-based DRL algorithm can refer to the reference \cite{haarnoja2018soft}. 

\subsection{Multi Omnidirectional Robot Formation Tracking Module}
This module aims to propose a novel HM-DRL algorithm for training the Omnidirectional robots to track the Ackermann-steering leader while
maintaining a certain predefined formation pattern. The core idea of this module is to decompose the complex, nonlinear formation tracking task into three control submodules: \textit{a high-precision physical feedforward controller}, \textit{a classical PD feedback controller}, and \textit{an adaptive RL residual controller}. 

It should be emphasized that the proposed HM-DRL control law for the $i$-th follower is computed in the leader's local frame $O_{L}$ and then transformed to the global frame $O_{W}$. The command velocity $\mathbf{v}_{local}^{(i)}$ is a superposition of three components
\begin{align}
\mathbf{v}_{local}^{(i)} = \mathbf{v}_{ff}^{(i)} + \mathbf{v}_{pd}^{(i)} + \mathbf{v}_{rl}^{(i)}
\end{align}
where $\mathbf{v}_{ff}^{(i)}$, $\mathbf{v}_{pd}^{(i)}$ and $\mathbf{v}_{rl}^{(i)}$ are the output of the three control submodules, respectively.

\subsubsection{Physical feedforward controller $\mathbf{v}_{ff}^{(i)}$} Serving as the ``brain'' and the foundation of the proposed HHy-PIDRL framework, this submodule is designed with a core objective that shifts from conventional error correction to active prediction. Specifically, it seeks to answer: "Given a theoretically error-free state, what should the follower's velocity be at the next moment to maintain the desired formation perfectly?" By precisely computing the dynamic influence of the leader's motion on the follower's target position, the controller generates an ideal velocity command. The proposed physical feedforward controller $\mathbf{v}_{ff}^{(i)}$  is defined as 
\begin{align}
\mathbf{v}_{ff}^{(i)} = \mathbf{v}_{lin}^{(i)} + \mathbf{v}_{tan}^{(i)} + \mathbf{v}_{cen}^{(i)}+\mathbf{v}_{euler}^{(i)}
\end{align}
where $\mathbf{v}_{lin}^{(i)}$ is the linear velocity compensation,  $\mathbf{v}_{tan}^{(i)}$ is the tangential velocity compensation, $\mathbf{v}_{cen}^{(i)}$ is the  centripetal acceleration compensation, and  $\mathbf{v}_{euler}^{(i)}$ is the  Euler acceleration compensation.  Below, we present the calculation steps for the four key parts. 

$\bullet$ Linear velocity compensation: $\mathbf{v}_{lin}^{(i)}$ is calculated to compensate for the leader's own translational speed. 
\begin{align}
\mathbf{v}_{lin}^{(i)}= 
\begin{bmatrix} 
v_{L,x}^{t} \cos(\theta_{L}^{t}) + v_{L,y}^{t} \sin(\theta_{L}^{t}) \\ 
-v_{L,x}^{t} \sin(\theta_{L}^{t}) + v_{L,y}^{t} \cos(\theta_{L}^{t}) 
\end{bmatrix}
\end{align}
where $v_{L,x}^{t}=v_{L}^{t}\mathrm{cos}(\theta_{L}^{t})$, $v_{L,y}^{t}=v_{L}^{t}\mathrm{sin}(\theta_{L}^{t})$.

$\bullet$ Tangential velocity compensation: The purpose of tangential velocity compensation $ \mathbf{v}_{tan}^{(i)}$ is to counteract the effects caused by the rotation of the leader itself. 
{\setlength\abovedisplayskip{1pt}
\setlength\belowdisplayskip{1pt}
\begin{align}
 \mathbf{v}_{tan}^{(i)} = 
\begin{bmatrix} 
-\omega_{L}^{t} Y^{t}_{i,\text{offset}} \\ 
\omega_{L}^{t}  X^{t}_{i,\text{offset}} 
\end{bmatrix}
\end{align}}
where $\omega_{L}^{t}$ is the angular speed of the leader robot, $Y_{t}^{i,\text{offset}}$ and $X_{t}^{i,\text{offset}}$  denote  the expected Y-coordinate and X-coordinate of the $i$th follower in $O_L$, respectively. The angular speed  $\omega_{L}^{t}$ can be calculated as 
{\setlength\abovedisplayskip{1pt}
\setlength\belowdisplayskip{1pt}
\begin{align}
\omega_{L}^{t}=\frac{(\Delta\theta_{\text{raw}}+\pi)(\mathrm{mod}(2\pi))-\pi}{\Delta t}
\end{align}}
where $\Delta\theta_{\text{raw}} = \theta_L(t) - \theta_L(t-\Delta t)$. 

$\bullet$ Centripetal acceleration compensation:  $\mathbf{v}_{cen}^{(i)}$ is given to  compensate for the centripetal acceleration required for followers to maintain their relative position when the leader turns. $\mathbf{v}_{cen}^{(i)}$ is calculated as 
{\setlength\abovedisplayskip{1pt}
\setlength\belowdisplayskip{1pt}
\begin{align}
\mathbf{v}_{cen}^{(i)}=
\begin{bmatrix} 
-\omega_{L}^{t^{2}} X_{i,\text{offset}}^{t}\Delta t \\ 
\omega_{L}^{t^{2}}  Y_{i,\text{offset}}^{t}\Delta t 
\end{bmatrix}
\end{align}}

$\bullet$ Euler acceleration compensation: $\mathbf{v}_{euler}^{(i)}$ is calculated to compensate for the Euler acceleration generated when the leader's angular velocity changes, and is given as
\begin{align}
\mathbf{v}_{euler}^{(i)}=
\begin{bmatrix} 
a_{L}^{t} Y_{i,\text{offset}}^{t}\Delta t \\ 
-a_{L}^{t}  X_{i,\text{offset}}^{t}\Delta t 
\end{bmatrix}
\end{align}
where $a_{L}^{t}=\dot{\omega}_{L}^{t}$. 

\subsubsection{PD feedback controller} The PD feedback controller is the ``corrector'' of proposed HHy-PIDRL framework. Its function is to passively respond to errors that have already occurred. It measures the error between the current actual position of the follower and the desired position, and generates a corrective velocity based on this deviation and its rate of change. The output of PD feedback controller is calculated as 
\begin{align}
V^{t}_{i,pd} = K_{p}  e_{i,\mathrm{pos}}^{t} + K_{d}  \dot{e}_{i,\mathrm{pos}}^{t}
\end{align} 
where $K_{p}$ and $K_{d}$ are the proportional gain and  derivative gain, respectively. In this paper, $K_{p}=5$, $K_{d}=1.5$. 
$e_{i,\mathrm{pos}}^{t}=P_{i,\mathrm{offset}}^{t}-P_{i,\mathrm{actual}}^{t}$, where 
$P_{i,\mathrm{offset}}^{t}=[X_{i,\mathrm{offset}}^{t},Y_{i,\mathrm{offset}}^{t}]^{T}$ is the desired position, $P_{i,\mathrm{actual}}^{t}=[X_{i,\mathrm{actual}}^{t},Y_{i,\mathrm{actual}}^{t}]^{T}$ is the actual position  in $O_L$. 

\subsubsection{Adaptive RL residual controller} Acting as the fine‑tuning and adaptive module of the proposed HHy‑PIDRL framework, the RL‑based residual controller is designed to learn and compensate for the residual error that persist after the combined action of the feedforward and PD controllers. Its role is to continuously refine the control output by adaptively correcting deviations that are not fully addressed by the preceding control stages. In the following, we detail the state space, action space, reward function, and network training procedure for each Omnidirectional folower. 

$\mathbf{State~Space}$: Assume the state space $\mathcal{S}_{i}^{t}$ of the $i$th follower robot is defined at time $t$, which is a high-dimensional continuous vector designed to provide all the necessary information for its decision-making. The state space $S_t^i$ is a 20-dimensional vector, and can be decomposed into the following five parts 
$\mathcal{S}_i^t = [ \mathcal{S}_{i,\text{error}}^t, \mathcal{S}_{\text{leader}}^{t}, \mathcal{S}_{i,\text{peers}}^t, \mathcal{S}_{i,\text{history}}^t, S_{\text{global}}^{t} ]^T $, where $\mathcal{S}_{i,\text{error}}^t$ denotes the error information of the follower itself, $\mathcal{S}_{\text{leader}}^{t}$ is the leader's dynamic information, $\mathcal{S}_{i,\text{peers}}^t$ denotes the relative information between each follower and its neighbours, $\mathcal{S}_{i,\text{history}}^t$ is the  follower's own historical errors, and $S_{\text{global}}^{t}$  represents the global formation error information. 

$\bullet$ $\mathcal{S}_{i,\text{error}}^t=[s_{1}^{t},s_{2}^{t}, s_{3}^{t}, s_{4}^{t}]^{T}$, $s_{i}^{t}$, $i=1,2,3, 4$ can be calculated as 
\begin{align}
s_{1}^{t}&=(X_{i,\text{offset}}^t - X_{i,\text{actual}}^t)/r_{\text{norm}}\\
s_{2}^{t}&=(Y_{i,\text{offset}}^t - Y_{i,\text{actual}}^t)/r_{\text{norm}}\\
s_{3}^{t}&=(v_{i,x,\text{actual}}^t - v_{L,x,\text{local}}^t)/v_{f,\text{max}}\\
s_{4}^{t}&=(v_{i,y,\text{actual}}^t - v_{L,y,\text{local}}^t)/v_{f,\text{max}}
\end{align}
where $[v_{i,x,\text{actual}}^t,v_{i,y,\text{actual}}^t]^{T}$ is the actual velocity of followers in $O_L$, $[v_{L,x,\text{local}}^t, v_{L,y,\text{local}}^t]^{T}$ is the projection of leader's velocity in $O_L$. $r_{\text{norm}}$ is the normalized radius,  $v_{f,\text{max}}$ is the maximum velocity of followers. 

$\bullet$ $\mathcal{S}_{\text{leader}}^{t}$ is considered to provide the current movement status of the leader to help followers predict future movement trends, and is defined as  $\mathcal{S}_{\text{leader}}^{t}$$=[v_{l}^{t}/v_{f,\text{max}}$,$\omega_{l}^{t}/\omega_{f,\text{max}}$, $a_{L,x}^{t}/a_{f,\text{max}}$, $a_{L,y}^{t}/a_{f,\text{max}}]^{T}$, where  $\omega_{f,\text{max}}$ and $a_{f,\text{max}}$ represent the maximum angular velocity and maximum acceleration of the followers, respectively. 

$\bullet$  $\mathcal{S}_{i,\text{peers}}^t=[s_{5}^{t},s_{6}^{t},s_{7}^{t},s_{8}^{t}]^{T}$,   $s_{i}^{t}$, $i=5,6,7, 8$ can be calculated as 
\begin{align}
s_{5}^{t}&=(x_{j}^{t} - x_{i}^{t})/r_{\text{norm}}\\
s_{6}^{t}&=(y_{j}^{t} - y_{i}^{t})/r_{\text{norm}}\\
s_{7}^{t}&=(v_{j,x,O_{W}}^t - v_{i,x,O_{W}}^t)/v_{f,\text{max}}\\
s_{8}^{t}&=(v_{j,y,O_{W}}^t - v_{i,y,O_{W}}^t)/v_{f,\text{max}}
\end{align}
where $x_{i}^{t}$, $x_{j}^{t}$,  $v_{i,x,O_{W}}^t$, and  $v_{i,y,O_{W}}^t$  denote the positions and velocities in $O_{W}$.  

$\bullet$  $\mathcal{S}_{i,\text{history}}^t=[s_{9}^{t}, s_{10}^{t}]^{T}$ is considered to provide error information from the past two steps to help the follower robots determine the trend of error changes.  $s_{9}^{t}$ and $s_{10}^{t}$ are calculated as 
\begin{align}
s_{9}^{t}&=|e_{i,\mathrm{pos}}^{t-2}|/r_{\text{norm}}\\
s_{10}^{t}&=|e_{i,\mathrm{pos}}^{t-1}|/r_{\text{norm}}
\end{align}

$\bullet$ $S_{\text{global}}^{t}=[s_{11}^{t},s_{12}^{t}]^{T}$ is given to provide the current average performance of the entire formation that can  help individuals adjust strategies to align with the overall. $s_{11}^{t}$ and $s_{12}^{t}$ are calculated as 
\begin{align}
s_{9}^{t}&=\bar{e}_{i,\mathrm{pos}}^{t}/r_{\text{norm}}\\
s_{10}^{t}&=\bar{e}_{i,\mathrm{dist}}^{t}/r_{\text{norm}}
\end{align}
where $\bar{e}_{i,\mathrm{pos}}^{t}$ and $\bar{e}_{i,\mathrm{dist}}^{t}$ represent the average position error and average spacing error, respectively. 

$\mathbf{Action~Space}$:  The action space of each  Omnidirectional follower robots  is defined as $A_{i}^{t}=[a_{i,x}^{t},a_{i,y}^{t}]^{T}$. Note that 
$[a_{i,x}^{t},a_{i,y}^{t}]^{T}$ is not the final speed command, but represents the original signal strength of the correction or adjustment that the follower robots expect to apply in the $X$ and $Y$ directions of the leader's local coordinate system  $O_{L}$. This obtained original signal is then multiplied by a scaling factor $a_{scaling}$  to obtain the final residual correction velocity $\mathbf{v}_{rl}^{(i)}$. 

$\mathbf{Reward~Function}$: In order to guide agents to learn high-precision, stable, and safe formation tracking behaviour, we design the reward function $\mathcal{R}_i^t$ for each follower robot at time step $t$. The total reward $\mathcal{R}_i^t$  is the sum of the following terms:
\begin{align}
\mathcal{R}_i^t &= \mathcal{R}_{i,\text{pos}}^{t}+ \mathcal{R}_{i,\text{spacing}}^{t}+\mathcal{R}_{i,\text{holding}}^{t}\nonumber\\
&~~~+ \mathcal{R}_{i,\text{progress}}^{t}+ \mathcal{R}_{\text{i,proximity}}^{t}+ \mathcal{R}_{i,\text{terminal}}^{t}
\end{align}

$\bullet$ Distance reward $\mathcal{R}_{i,\text{pos}}^{t}$:  Distance reward is designed to encourage follower robots to approach their corresponding desired positions, and is given by 
\begin{align}
R_{i,\text{pos}}^{t} = C_{\text{pos}}  e^{-k_{\text{pos}}  e_{i,\text{pos}}^{t}}
\end{align}
where $C_{\text{pos}}=8$, $k_{\text{pos}}=2$, $e_{i,\text{pos}}^{t}$ represents the Euclidean distance between $i$th Omnidirectional follower at the current moment and its desired position. 

$\bullet$ Spacing reward $\mathcal{R}_{i,\text{spacing}}^{t}$:   Spacing reward is designed to encourage each follower robot to maintain the correct relative distance with their neighbours  to preserve the formation structure. $\mathcal{R}_{i,\text{spacing}}^{t}$ is considered as 
\begin{align}
R_{i,\text{spacing}}^{t} = C_{\text{spacing}} e^{-k_{\text{spacing}}  \bar{e}_{i,\text{dist}}^{t}}
\end{align}
where  $C_{\text{spacing}}=4$, $k_{\text{spacing}}=3$,  $\bar{e}_{i,\text{dist}}^{t}$ represents the average spacing error at the current moment between $i$th follower and all its neighbours.

$\bullet$ Holding reward $\mathcal{R}_{i,\text{holding}}^{t}$: Holding reward is designed to give an additional reward that increases sharply as the error decreases, encouraging the agent to achieve and maintain high-precision hovering when the follower is very close to the desired position. $\mathcal{R}_{i,\text{holding}}^{t}$ is designed as 
$$ 
R_{i,\text{holding}}^{t}= \begin{cases} C_{\text{holding}} e^{-k_{\text{holding}} (e_{i,\text{pos}}^{t})^2} & \text{if } e_{i,\text{pos}}^{t}< \tau_{\text{holding}} \\ 0 & \text{otherwise} \end{cases}$$
where $C_{\text{holding}}=6$, $k_{\text{holding}}=10$, $\tau_{\text{holding}}=0.3$.

$\bullet$ Progress reward $\mathcal{R}_{i,\text{progress}}^{t}$: Progress reward is designed to reward each follower robot for taking actions that reduce the position error, encouraging it to move in the correct direction. $\mathcal{R}_{i,\text{progress}}^{t}$ is designed as 
\begin{align}
R_{i,\text{progress}}^{t}= C_{\text{progress}}\max(0, e_{i,\text{pos}}^{t-1} - e_{i,\text{pos}}^{t})
\end{align}
where $C_{\text{progress}}=3.0$. 

$\bullet$  Soft collision penalty $\mathcal{R}_{\text{i,proximity}}^{t}$: Soft collision penalty  is divided into two parts: the proximity penalty for the leader $R_{iL}^{t}$ and the proximity penalty for other follower robots $R_{ij}^{t}$. $\mathcal{R}_{\text{i,proximity}}^{t}$ is given as 
\begin{align}
R_{i,\text{proximity}}^{t}=R_{iL}^{t}+\sum_{j=1, j \neq i}^{N} R_{ij}^{t}, 
\end{align}
where $N$ is the number of follower robots. $R_{iL}^{t}$ is calculated as 
\begin{align}
R_{iL}^{t}=\begin{cases}
-C_{\text{prox}}\left(1 - \frac{d_{iL}^{t}}{d_{\text{safe,L}}}\right)^2 & \text{if } d_{iL}^{t}< d_{\text{safe,L}} \\
0 & \text{if } d_{iL}^{t}\ge d_{\text{safe,L}}
\end{cases}
\end{align}
where $C_{\text{prox}}=6.0$, $d_{iL}^{t}$ denotes the Euclidean distance between the $i$th follower robot and the leader. $d_{\text{safe,L}}$ is the safe distance threshold. $R_{ij}^{t}$ is calculated as 
\begin{align}
R_{ij}^{t}=\begin{cases}
-C_{\text{prox}}\left(1 - \frac{d_{ij}^{t}}{d_{\text{safe,F}}}\right)^2 & \text{if } d_{ij}^{t}< d_{\text{safe,F}} \\
0 & \text{if } d_{ij}^{t}\ge d_{\text{safe,F}}
\end{cases}
\end{align}
where $d_{ij}^{t}$ and $d_{\text{safe,F}}$ is the Euclidean distance and safe distance threshold between follower $i$ and follower $j$. 

$\mathbf{Network~Trainning}$: Similarly, the SAC-based DRL algorithm is employed to train all Omnidirectional followers, enabling them to track the leader while maintaining a specified formation geometry. To enhance sample efficiency, a prioritized experience peplay (PER) buffer is incorporated. Unlike standard uniform sampling, PER prioritizes transitions according to their estimated learning potential, typically measured by temporal-difference (TD) error so that experiences with higher prediction inaccuracy and larger TD error are replayed more frequently. This mechanism directs the learning process toward the most informative samples, thereby accelerating training convergence and improving the final control policy. 
The overall formation control policy network for the Omnidirectional followers, based on this HM-DRL approach, is summarized in Algorithm \textcolor[rgb]{0.00,0.07,1.00}{1}.

\begin{algorithm}\label{T1}
\scriptsize  
\caption{\footnotesize Hybrid Model and DRL (HM-DRL) based Formation Control Policy Networks For Omnidirectional Followers}
\begin{algorithmic}[1]
\vspace{-1pt} 
\State $\mathbf{Input}$: The trained policy network $\pi^*_{L}$ for Ackermann-steering leader, formation control environment with $3$ Omnidirectional followers and an  Ackermann-steering leader, total number of episodes $(10000)$, max steps per episode $(300)$, batch size $(128)$, and soft update coefficient (0.005).

\State $\mathbf{Initialization}$:  Initialize the random initial position $P_{L}^{0}$ of the leader in the environmental map $O_{W}$. Initialize  initial positions of all the Omnidirectional followers. Initialize actor network $\pi_{\phi_{i}}$ with parameters $\phi_{i}$. Initialize critic networks $Q_{\theta_{i,1}}$, $Q_{\theta_{i,2}}$ with parameters $\theta_{i,1}$, $\theta_{i,2}$. Initialize target critics with $\bar{\theta}_{i,j}\leftarrow \theta_{i,j}$ for $j=1,2$. Initialize Entropy coefficient $\alpha_{i}\leftarrow 1.0$. 
Initialize the PER pool $D_{i}^{f}$ with capacity  $\mathcal{D}_{i}^{cap}$ and max priority  $p_{max} \leftarrow 1.0$. 
\For{each robot $i$}
    \For{each episode}
         \State Get the initial positions of the Ackermann-steering leader and all Omnidirectional followers. 
        \For{each step}
            \State Sample the leader's action $A^{t}_{L}\sim \pi^*_{L}(S^t_L)$ according to the observation $S^t_L$ at current time $t$.  
            Calculate and obtain the action $A_{i}^{t}$ for each Omnidirectional follower according to the observation $\mathcal{S}_i^t$  at current time $t$. 
            \State Update the kinematic model (\ref{E2})  and get the new state vector $\mathcal{S}_i^{t+1}$  and calculate $\mathcal{R}_i^t $. 
            \State Store experience into separate PER pools $D_{i}^{f} \leftarrow D_{i}^{f} \cup (\mathcal{S}_i^t, A_{i}^{t}, \mathcal{R}_i^t, \mathcal{S}_i^{t+1})$.
            \State $\mathbf{Prioritized~Sampling }$: Compute probabilities for all $k \in \mathcal{D}_i$: $P(k) \leftarrow \frac{p_k^\nu}{\sum_j p_j^\nu}$. Sample batch indices $\mathcal{B}$ based on $P(k)$. Compute importance sampling weights for $k \in \mathcal{B}$: $w_k \leftarrow ( \frac{1}{|\mathcal{D}_i| \cdot P(k)} )^\beta$. $\nu$ is the hyperparameter of priority usage,  $\beta$ is an importance sampling scalar that counteracts the non-uniform sampling deviations. 

            \State $\mathbf{Critic~Update}$: Sample next actions $\tilde{A}'_{i} \sim \pi_{\phi_i}(\cdot | \mathcal{S}_{i,k+1})$.
                Compute Target Q-values: $y_{i,k} \leftarrow R_{i,k} + \gamma (1 - d_{i,k}) \left( \min_{j=1,2} Q_{\bar{\theta}_{i,j}}(\mathcal{S}_{i,k+1}, \tilde{A}'_{i}) - \alpha_i \log \pi_{\phi_i}(\tilde{A}'_{i} | \mathcal{S}_{i,k+1}) \right)$, where  $\gamma$ is the discount factor,  $Q_{\bar{\theta}_{i,j}}$ is the parameter of target critics network, and $\alpha_i$ is the temperature coefficient.  
                Compute Critic Loss: $L_Q(\theta_i) = \frac{1}{B} \sum_{k \in Batch} w_k \left( Q_{\theta_{i,j}}(S_{i,k}, A_{i,k}) - y_{i,k} \right)^2$. Update $\theta_{i,j}$ via gradient descent. 
        
            \State $\mathbf{Priorities~Update}$: Calculate TD-error for PER: $\delta_k = |y_{i,k} - Q_{\theta_{i,1}}(S_{i,k}, A_{i,k})|$. Update priorities in $\mathcal{D}_i^{f}$: $p_k \leftarrow (|\delta_k| + \epsilon)$, $\epsilon$ is a very small positive scalar.
            
            \State $\mathbf{Actor~Update}$: Update $\phi_i$ by maximizing expected return and entropy: 
            $\phi_i \leftarrow \phi_i - \lambda_\pi \cdot \nabla_{\phi_i} ( \frac{1}{B} \sum_{k \in \mathcal{B}} ( \alpha_i \log \pi_{\phi_i}(A_{i,k} | \mathcal{S}_{i,k}) - \min_{j=1,2} Q_{\theta_{i,j}}(\mathcal{S}_{i,k},A_{i,k}) ))$, 
            $\lambda_\pi$ is the learning rate. 
            \State $\mathbf{Temperature~Auto-Tuning}$: Update $\alpha_i$ to match target entropy $\bar{\mathcal{H}}$: 
            $\alpha_i \leftarrow \alpha_i - \lambda_\alpha \cdot \nabla_{\alpha_i} (- \frac{1}{B} \sum_{k \in \mathcal{B}} \alpha_i \left( \log \pi_{\phi_i}(A_{i,k} | \mathcal{S}_{i,k}) + \bar{\mathcal{H}} \right))$, $\lambda_\alpha$ is the learning rate. 
        \EndFor
        \State $\mathbf{Soft~Target~Update}$: $\bar{\theta}_{i,j} \leftarrow \tau \theta_{i,j} + (1 - \tau) \bar{\theta}_{i,j}, \quad j=1,2$. $\tau$ is the soft update coefficient. 
       
    \EndFor
\EndFor

\State \textbf{Return:} Optimized parameters $\phi^*_i, \, i \in \{1, 2, ..., n\}$.
\vspace{-2pt} 
\end{algorithmic}
\end{algorithm}



\section{Simulation And Experiment Results}\label{S4}
Simulation and experiment results are presented to validate the effectiveness of the proposed HHy-PIDRL framework. The results obtained from the \textit{single Ackermann-steering robot navigation module} and  \textit{multi Omnidirectional robot formation tracking module} are conducted on a server with an Intel Xeon Platinum 8352S processor and an NVIDIA GeForce RTX 4090 GPU. 
All simulations and experiments are performed in a two‑dimensional plane defined within the region  $[0\mathrm{m},20\mathrm{m}]\times [0\mathrm{m},20\mathrm{m}]$, $\mathrm{m}$ is the unit of length. All trained networks in this paper are constructed as standard multi-layer perceptrons (MLPs), each uniformly comprising two hidden layers of $256$ neurons with the ReLU activation function.

\begin{table} 
\caption{Key Hyperparameters For training The Leader} 
 \centering  
\scalebox{0.8}{ 
\begin{tabular}{ccc}
   \toprule
Parameters & Value  \\
   \midrule
Entropy Temperature Coefficient $\alpha$ & $0.3$  \\
Discount Factor $\gamma$ &   $0.99$\\
Soft Update Rate $\tau$  &  $0.005$\\
Critic Learning Rate &  $10^{-3}$ \\
Actor Learning Rate &   $10^{-3}$ \\
Hidden Dimension & $256$\\
Batch Size        & $256$\\
Replay Buffer Size&$800000$ \\
   Maximum Steps & $300$\\ 
\bottomrule
\end{tabular}
}
\label{table1}
\end{table} 

\vspace{-1.0em}
\subsection{Single Ackermann-Steering Robot Navigation Module}
Compared with the existing results \cite{li2024distributed}, \cite{du2025scalable}, \cite{goeckner2024graph} that adopt particle-based models, this paper employs a actual model for training. For the leader robot, the Ackermann-steering model (\ref{E1}) is used. Some physical parameters of the Ackermann-steering leader robot are shown in TABLE \ref{table2}.

\begin{table} 
\vspace{-2.0em}
\caption{physical parameters of Ackermann-steering leader} 
 \centering  
 \scalebox{0.8}{ 
\begin{tabular}{ccc}
   \toprule
Parameters & Value  \\
   \midrule
Length of The Robot  & $0.7\mathrm{m}$  \\
Width of The Robot  &   $0.4\mathrm{m}$\\
Length of The Wheels   &  $0.1\mathrm{m}$\\
Width of The Wheels  &  $0.08\mathrm{m}$ \\
Maximum Steering Angle $\phi_{\mathrm{max}}$ & $\pi/6$\\
Maximum Speed  $v_{L_{\mathrm{max}}}$ &   $3\mathrm{m}/s$ \\
Change Rate of Maximum Steering Angle  & $\pi/3$\\
\bottomrule
\end{tabular}
}
\label{table2}
\end{table}

The objective of this module is to train the Ackermann‑steering leader robot to navigate autonomously to a target position, defined as a circular region of radius $0.5\mathrm{m}$.
During training, both the initial and target positions are randomized in each episode, with a minimum separation of $8\mathrm{m}$ enforced to ensure meaningful learning
Each episode employs a time step of $\Delta t=0.1s$ and terminates when the robot either reaches the maximum allowed step count of $300$ or collides with the environment boundary.
The leader is trained using the SAC-based DRL algorithm, with key network hyperparameters listed in TABLE \ref{table1}. 
Training is conducted over $50000$ episodes, with the environment reset at the start of each episode. The final trained model is saved for subsequent testing and evaluation.

\begin{figure}[!htp]
\centering
\includegraphics[scale=0.3]{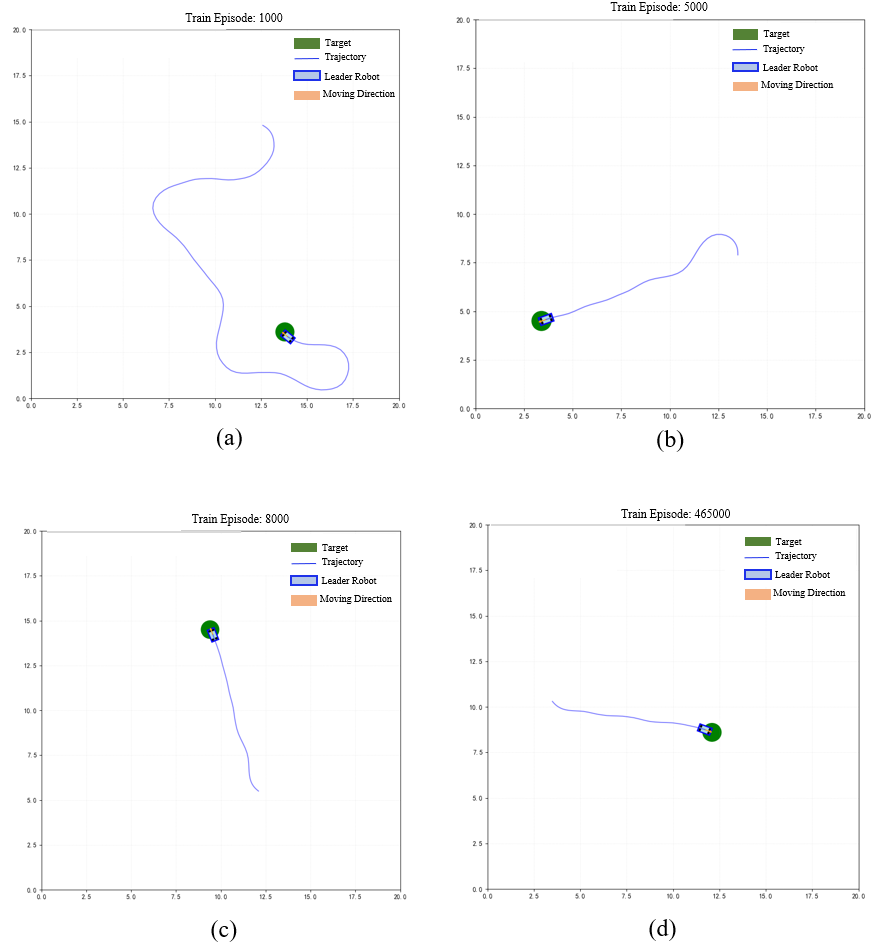}
\caption{Visualized results of training at four different  episodes.}
\label{fig333}
\end{figure}

During the training process, the training results are also visualized at regular intervals.  
Specifically, Fig. \ref{fig333} presents the training performance at training  episodes $1000$, $5000$, $8000$, and $46500$. At $1000$th episode, it can be seen that although the leader robot can finally reach the target position, it takes a relatively long time. As the number of training episode increases, it can be seen that the trajectory of the leader robot reaching the target position is relatively smooth and takes less time.

\begin{figure}[!htp]
\centering
\includegraphics[scale=0.2]{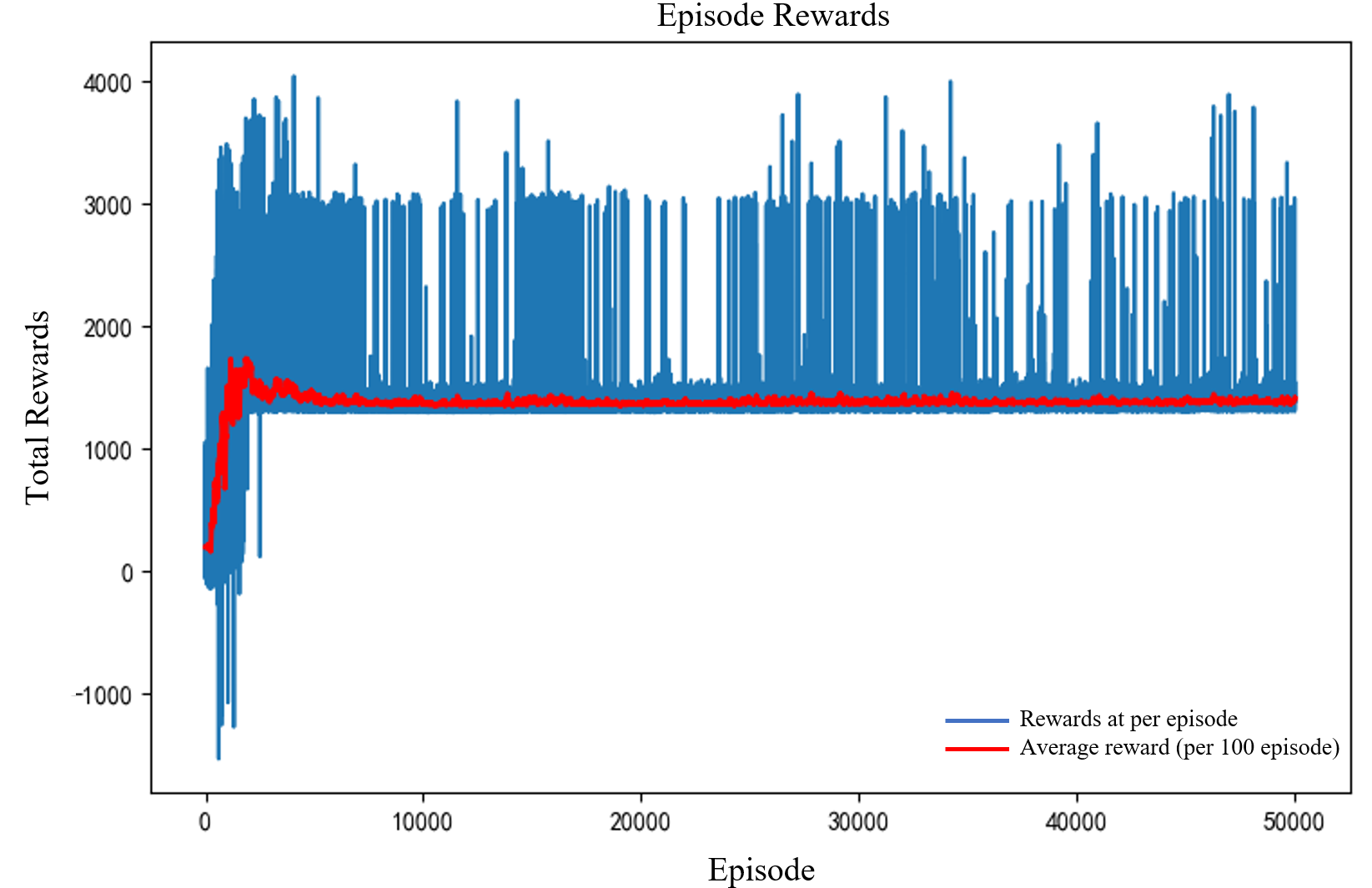}
\caption{Reward curves obtained during the exploratory training process of the leader robot.  The blue curve corresponds to the episodic reward, while the red curve represents a sliding‑window average computed over every $100$ episodes. } 
\label{fig444}
\end{figure}
In Fig. \ref{fig444}, the reward curves  of the leader robot during the process of being trained to explore are presented.  The blue curve corresponds to the raw reward data, which exhibits substantial fluctuations as it records the reward. 

The red curve represents the sliding average reward, computed over a window of 100 consecutive episodes.  This metric averages the reward values from the most recent 100 episodes and is plotted to generate a smoothed curve that clearly reflects the overall training trend. As shown, the average reward steadily increases and eventually converges, indicating that the leader robot has successfully learned a stable and efficient policy for reaching randomly assigned target positions. 

\begin{figure}[!htp]
\centering
\includegraphics[scale=0.2]{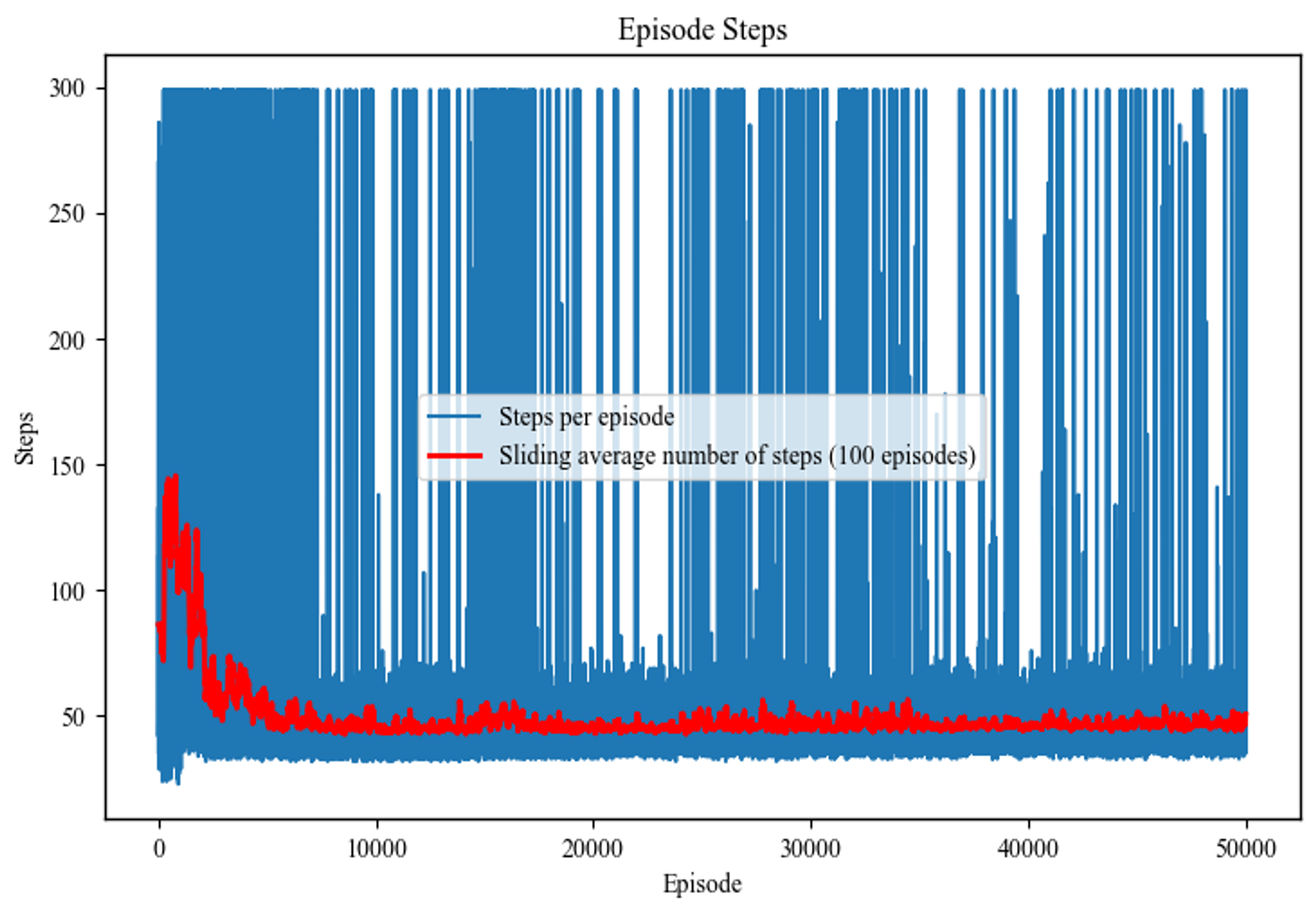}
\caption{Step numbers of the leader reaching the target positions. The blue line represents the number of steps taken by the leader to reach the positions in each episode, and the red line represents the average number of steps taken by the leader to reach the target position every $100$ episodes.  } 
\label{fig555}
\end{figure}

\begin{figure}[!htp]
\centering
\includegraphics[scale=0.2]{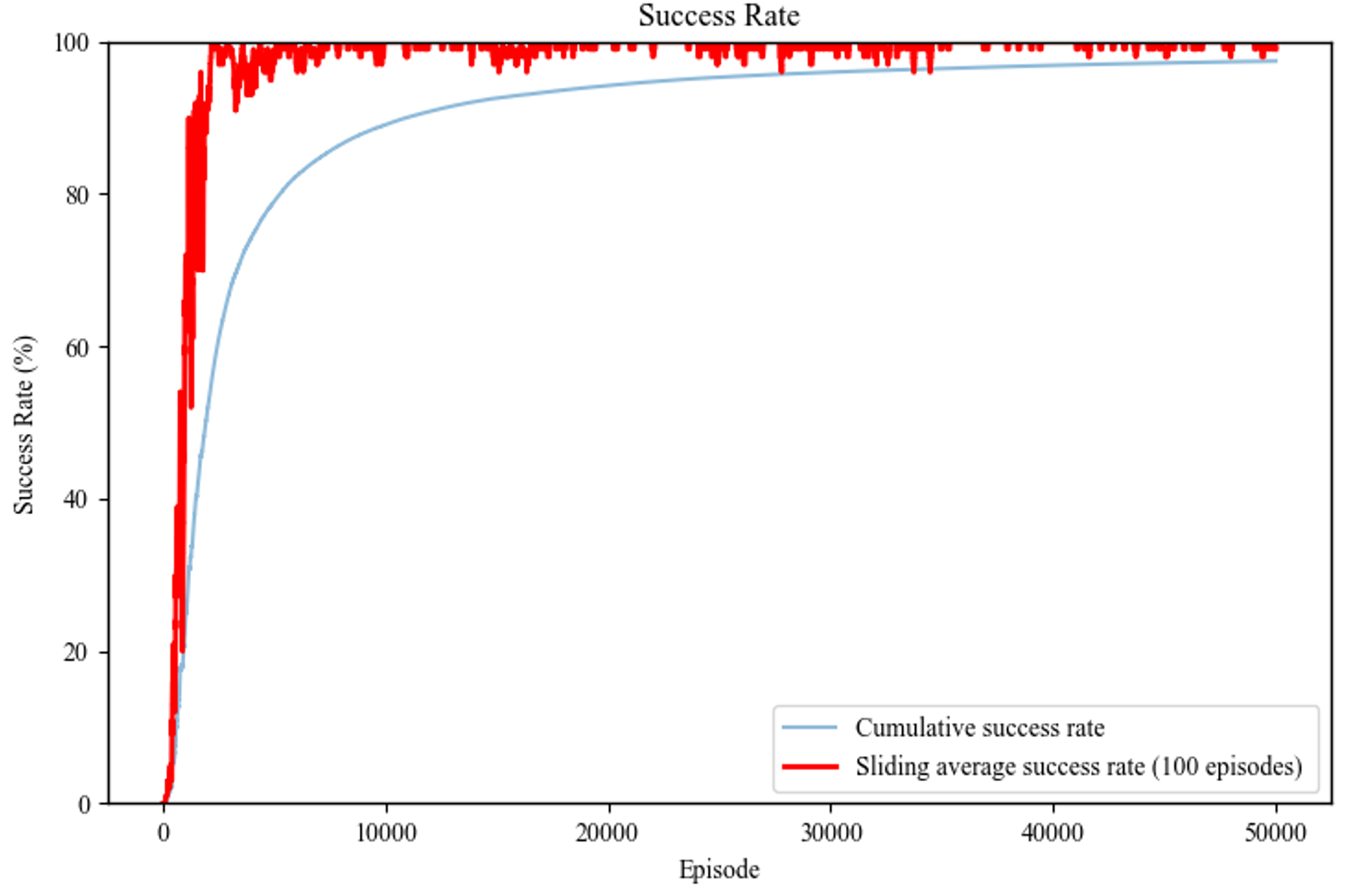}
\caption{Success rate of reaching the target  positions. The blue line represents the cumulative success rate, and the red line denotes the sliding average success rate every $100$  episodes. } 
\label{fig666}
\end{figure}

In Fig. \ref{fig555}, we give the numbers of steps taken by the leader robot to reach the target position, the blue line  represents the number of steps the leader takes to reach the target position in each training episode, the red line represents the average number of steps the leader takes to reach the target position every $100$ episodes. As we can see from the red line, the leader's steps to reach the target position stabilize to about $50$ steps,  which shows that the leader has learned a strategy that takes the shortest path and the least time to reach the target position. In Fig. \ref{fig666}, the success rate of reaching the target position is shown, we can see the  success rate is gradually approaching $100\%$. 

Next, to validate the performance of the trained model,  the leader robot’s initial and target positions are randomly generated.  The trained autonomous navigation model for the leader was then loaded, and $20$ independent test episodes are conducted. The experimental outcomes were statistically analyzed in terms of the average cumulative reward, the average number of steps per episode, and the average success rate of reaching the target position across all test episodes.
The statistical results are shown in TABLE. \ref{table3}. It can seen from TABLE. \ref{table3} that the average reward, average step, and average success rate  are $1390.82$, $48.1$, and $100\%$, respectively. It can be seen from the statistical results that the average values obtained are consistent with the results obtained by training.

\begin{table}[!htp]
\caption{statistical results} 
 \centering  
\begin{tabular}{ccc}
   \toprule
Test Results & Value  \\
   \midrule
Average Reward  & $1390.82$  \\
Average Step  &   $48.1$\\
Average Success Rate  &  $100\%$\\
\bottomrule
\end{tabular}
\label{table3}
\end{table} 
\vspace{-0.1em}

\subsection{Multi Omnidirectional Robot Formation Tracking Module}
The objective of this module is to verify the feasibility of the proposed HHy-PIDRL framework that can guide the Omnidirectional followers to tracking the leader while maintain specific formation structure. In order to ensure the authenticity and reliability of the results obtained, the followers also considered the actual kinematic model (\ref{E2}). Some physical parameters of the Omnidirectional followers are shown in TABLE \ref{table4}. 

In order to achieve stable and safe formation and tracking control, safe distance threshold  is set  as $d_{\text{safe,L}}=1.8\mathrm{m}$,  $d_{\text{safe,F}}=2.25\mathrm{m}$. In the paper, the triangular and linear formations are considered to verify the feasibility of the proposed HHy-PIDRL framework. 

\begin{figure}[!htp]
\centering
\includegraphics[scale=0.25]{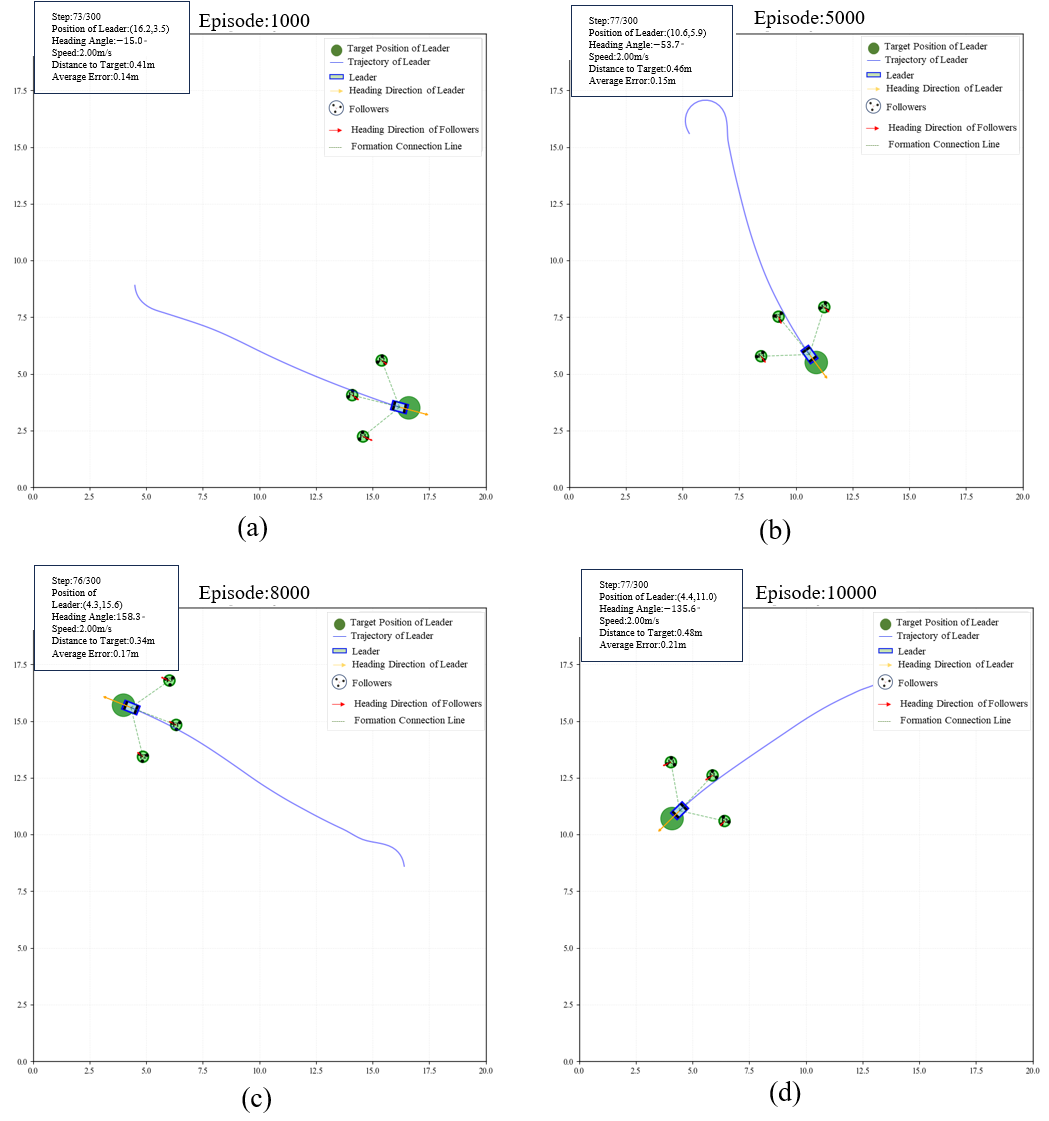}
\caption{Training results of different episodes. A stable policy network is trained for each follower to track the leader while maintain a  triangular formation.} 
\label{fig777}
\end{figure}

$\mathbf{Triangular~Formation}$ (shown in Fig. \ref{fig1}(c)): Three Omnidirectional followers track the leader while maintain the triangular formation. Follower $0$ is located 2m directly behind the leader, follower $1$ and follower $2$ are respectively located 2m behind the left and right of follower 0, and the angles with the follower are $60$ degrees and $-60$ degrees, respectively. It should be emphasized that the leader's strategy model here has been trained and its purpose is to navigate to a random target position. Here, each follower adopts an independent learning model and has its own network structure and priority experience pool. Then, SAC-based RL algorithm is design to train the Omnidirectional followers, and some specific network training hyperparameters are shown in TABLE \ref{table5}. The training episode is  $10000$. In Fig. \ref{fig777}, the training results of $1000$, $5000$, $8000$, $10000$ episodes are presented. From Fig. \ref{fig777}, as the number of training rounds increases, followers gradually improve their ability to track the leader in a triangular formation, which implies that a stable policy network is trained for each follower.

\begin{table} 
\caption{physical parameters of Omnidirectional followers} 
 \centering  
 \scalebox{0.8}{ 
\begin{tabular}{ccc}
   \toprule
Parameters & Value  \\
   \midrule
Radius of The Robot  & $0.24\mathrm{m}$  \\
Radius of The Wheels  &   $0.06\mathrm{m}$\\
Distance From Wheel to Center   &  $0.192\mathrm{m}$\\
Angle Distribution of Three Wheels  &  $(0,2\pi/3,4\pi/3)$ \\
Maximum Speed  $v_{f,{\mathrm{max}}}$ &   $5.4\mathrm{m}/s$ \\
\bottomrule
\end{tabular}
}
\label{table4}
\end{table}

\begin{table} 
\caption{Key Hyperparameters For training The Followers} 
 \centering  
 \scalebox{0.8}{ 
\begin{tabular}{ccc}
   \toprule
Parameters & Value  \\
   \midrule
Initial Temperature Coefficient $\alpha$ & $1$  \\
Discount Factor $\gamma$ &   $0.99$\\
Soft Update Rate $\tau$  &  $0.005$\\
Critic Learning Rate &  $3\times10^{-4}$ \\
Actor Learning Rate &   $3\times10^{-5}$ \\
Learning Rate $\beta$ & $0.0003$\\
Hidden Dimension & $256$\\
Batch Size        & $128$\\
Replay Buffer Size&$800000$ \\
   Maximum Step & $300$\\ 
Target Entropy   &  $-1.5$\\
Maximum Episode & $10000$\\
\bottomrule
\end{tabular}
}
\label{table5}
\end{table}

\begin{figure}[!htp]
\centering
\includegraphics[scale=0.23]{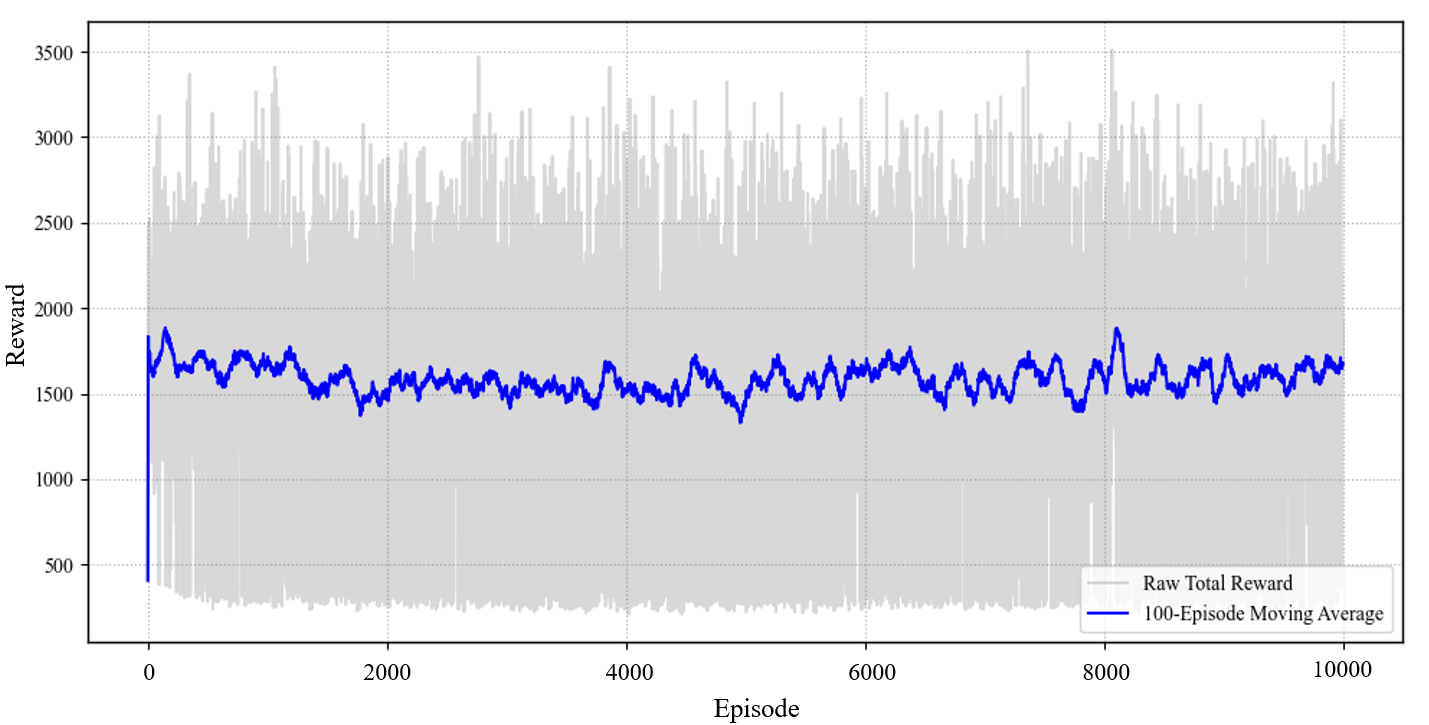}
\caption{Rewards per episode and average rewards over the last $100$ episodes.} 
\label{fig888}
\end{figure}

\begin{figure}[!htp]
\centering
\includegraphics[scale=0.3]{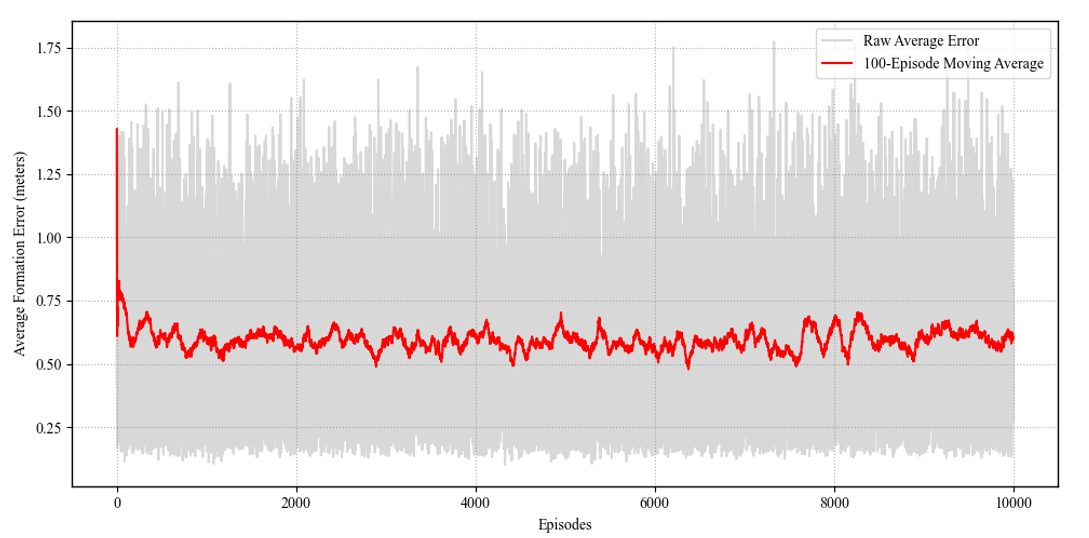}
\caption{Formation error per episode and average formation error over the last $100$ episodes. } 
\label{fig999}
\end{figure}

In Fig. \ref{fig888}, the rewards per episode and average rewards over the last $100$ episodes are shown.  It can be observed that with an increase in the number of training episodes, the moving average reward converges progressively, which indicates that the policy networks of the follower robots have been effectively trained. 
In Fig. \ref{fig999}, the formation error per episode and average formation error over the last $100$ episodes are shown. These results show  that the average formation error gradually converges, fluctuating around $0.6m$, which implies  the formation structure gradually stabilized as the number of training episodes increase. In this paper, the average position error is considered as  the average formation error, and is defined as
\begin{align}\label{E35}
\text{AvgPosErr}=\frac{1}{N_{\mathrm{e}}}\sum_{k=1}^{N_{\mathrm{e}}}\bigg(\frac{1}{T_{k}}\sum_{m=1}^{T_{k}}
\big(\frac{1}{N_{f}}\sum_{i=1}^{N_{f}}PosErr_{i,m,k}\big)\bigg)
\end{align}
where $N_{\mathrm{e}}$ is total number of test episodes, $k$ is the index of episodes, $m$ is the index of steps, $i$ is the index of followers, 
$T_{k}$ the  total steps of $k$-th episode, $N_{f}$ is the number of follower. $\text{PosErr}_{i,t,e} = \sqrt{(x_i - x_{i,\text{d}})^2 + (y_i - y_{i,\text{d}})^2}$, $(x_i,y_i)$ is actual position of follower $i$ in $O_{W}$, $(x_{i,\text{d}}, y_{i,\text{d}})$ is the desired position.

In order to test the trained policy networks for each follower, we define three metrics: average success rate,  average position error and average spacing error. Then,  $20$-episodes tests are conducted  within the same environmental map. 

\subsubsection{Average~Success~Rate} Percentage of successful numbers of episodes  where followers tracked the leader in a triangular formation within $20$ episodes relative to the total number of test episodes. Average success rate can be defined as $\frac{\sum N_{\mathrm{s}}}{N_{\mathrm{e}}}\times 100\%$, where $N_{\mathrm{s}}$ is the total number of episodes recorded  as “success.” It should be noted that the “success” is recorded  only when all three conditions are met simultaneously: 1) The followers track the leader to the target point; 2) No collisions occur between the leader and follower, nor between any followers; 3) The average position error for the last $30\%$ of steps in each round is less than 0.5m, and the average distance error is less than 0.5m.

\subsubsection{Average Position Error}  Please refer to (\ref{E35}).

\subsubsection{Average Spacing Error} It is an metric for measuring the internal structural stability of  formation control system. Average spacing error is defined as $\text{AvgDistErr}=\frac{1}{N_{\mathrm{e}}}\sum_{k=1}^{N_{\mathrm{e}}}\big(\frac{1}{T_{k}}\sum_{m=1}^{T_{k}}
\big(\frac{1}{N_{f}}\sum_{i=1}^{N_{f}}DistErr_{i,m,k}\big)\big)$, where $DistErr_{i,m,k}= \frac{1}{N_f - 1} \sum_{j=1, j \neq i}^{N_f} \text{Err}_{i,j}$,  $\text{Err}_{i,j}$  denotes  the Euclidean distance between two followers $i$ and $j$.

\begin{figure}[!htp]
\centering
\includegraphics[scale=0.4]{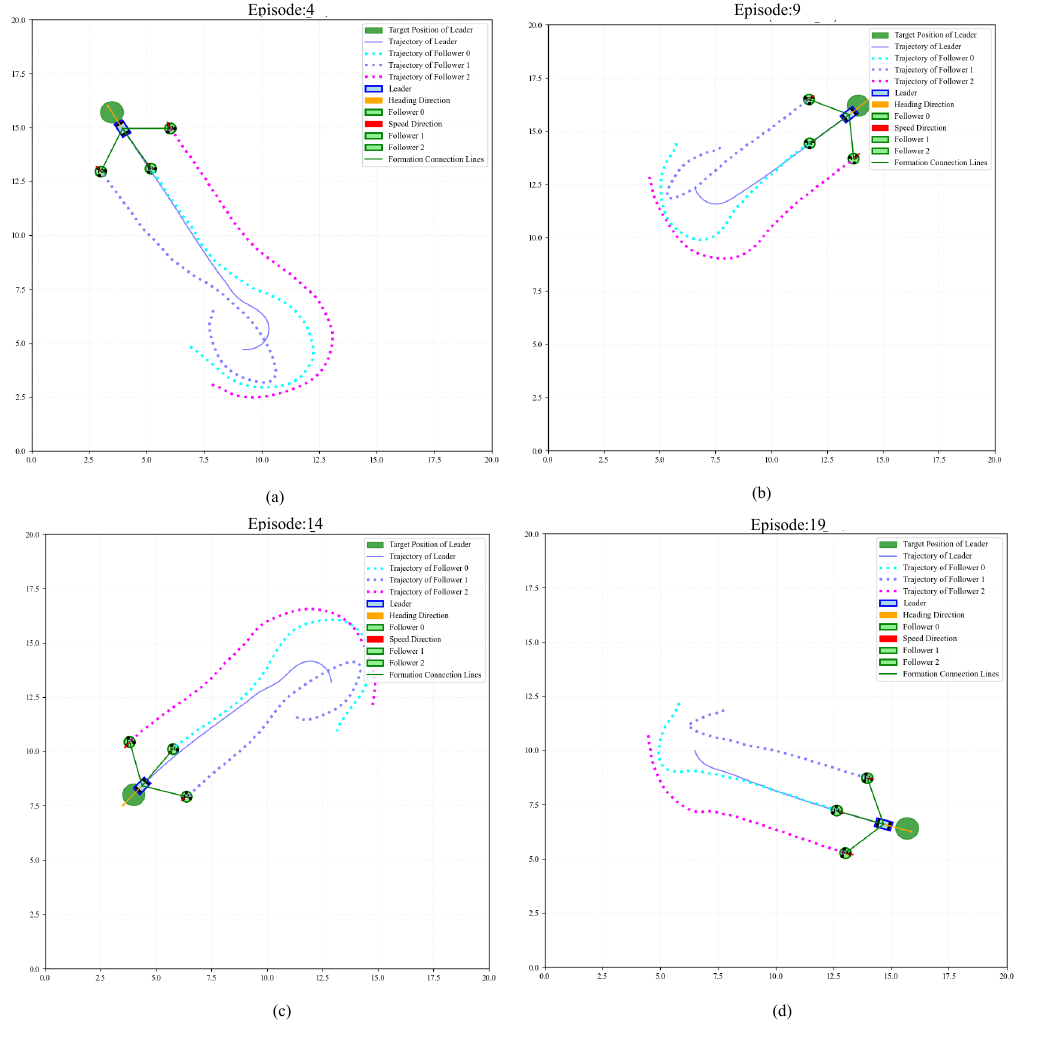}
\caption{Test experiments conducted in different episodes.} 
\label{fig1999}
\end{figure}

The statistical results of $20$-episodes tests are shown in TABLE \ref{table6}. We can see that the resulting policy networks ensure that the average position error and average spacing error of the triangular formation remain within a very small range, which directly demonstrates the feasibility of the obtained policy networks. Furthermore, in Fig. \ref{fig1999}, the test experiments conducted in different episodes are shown. As shown in Fig. \ref{fig1999}, the trained policy network ensures that followers track the leader in a specific triangular formation.

\begin{table}[!htp]
\caption{statistical results} 
 \centering  
\begin{tabular}{ccc}
   \toprule
Test Results & Value  \\
   \midrule
Average Position Error  &   $0.191m$\\
Average Success Rate  &  $100\%$\\
Average Spacing Error & $0.203m$\\
\bottomrule
\end{tabular}
\label{table6}
\end{table} 

\begin{figure}[!htp]
\centering
\includegraphics[scale=0.36]{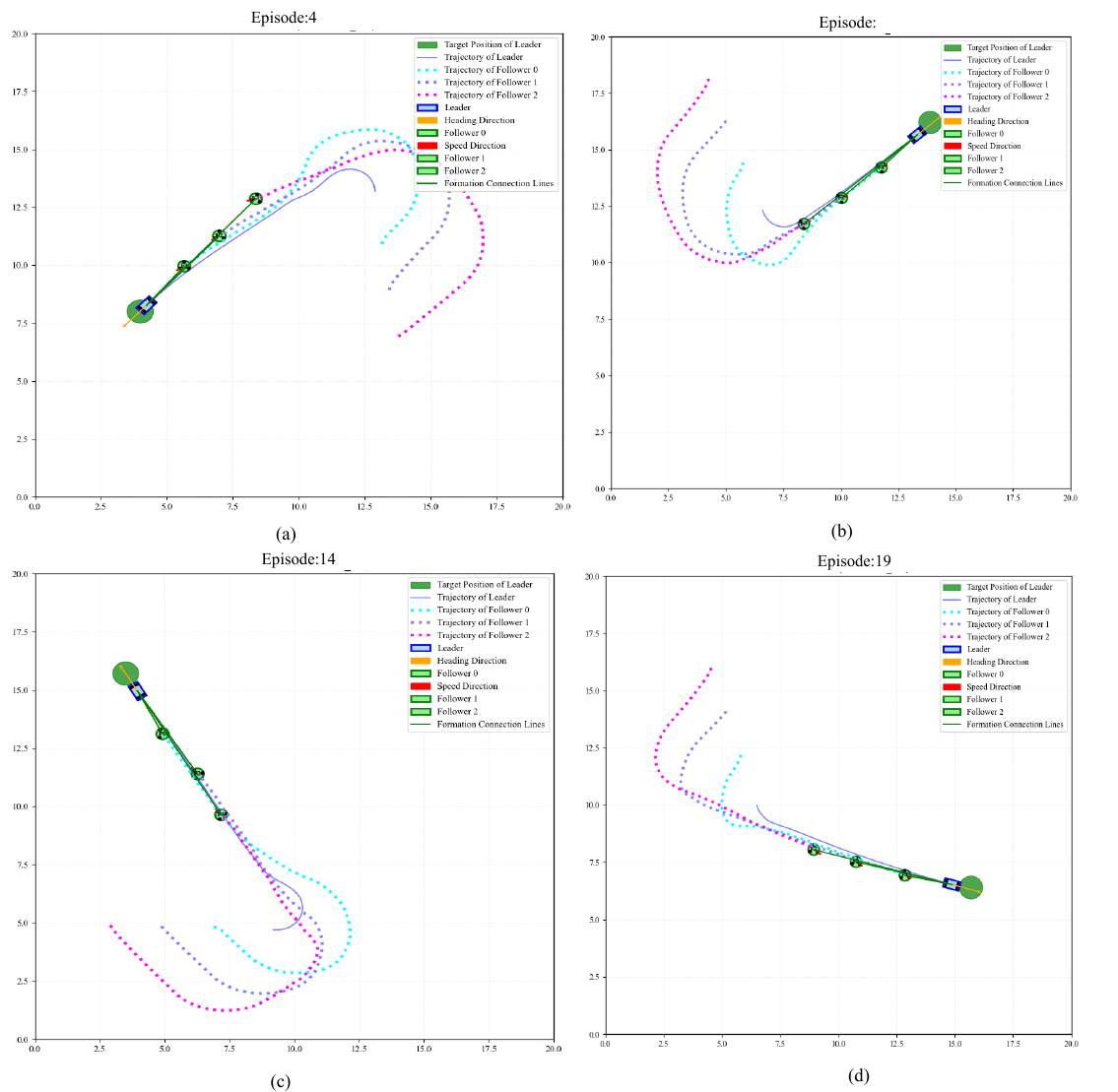}
\caption{Test experiments conducted in different episodes.} 
\label{fig11}
\end{figure}

$\mathbf{Vertical~Line~Formation}$ (shown in Fig. \ref{fig1}(d)): To better demonstrate the scalability of the trained policy networks, the networks are additionally deployed for the implementation of line formation. Fig. \ref{fig11} presents the corresponding test results across different episodes.
As illustrated in Fig. \ref{fig11}, the Omnidirectional follower robots gradually converge into a vertical line formation and achieve accurate tracking of the leader robot throughout the navigation process toward the target position.

\begin{figure*}
\centering
\includegraphics[scale=0.5]{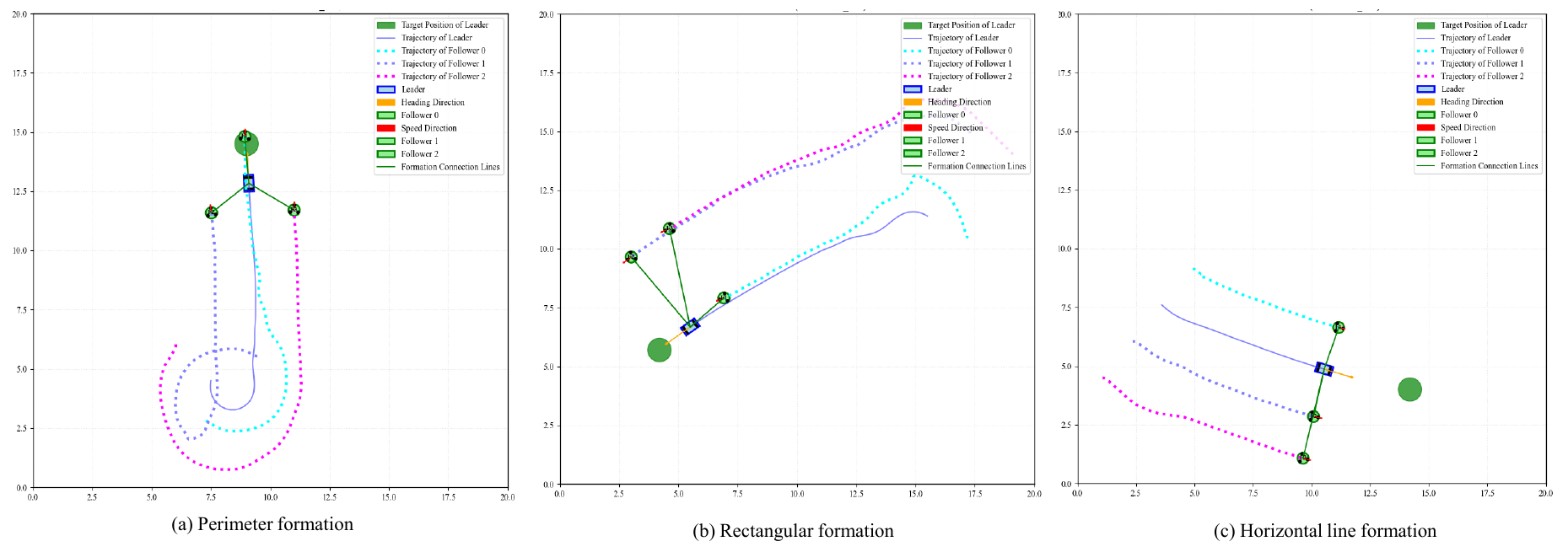}
\caption{Three common scalable formation patterns. }
\label{fig12}
\end{figure*}

It should be noted that the policy networks trained in this paper can be applied not only to triangular and linear formations but also to square formations, rectangular formations, and others, provided the formation structure is properly designed. Owing to space limitations,   three representative scalable formation patterns  are shown in Fig. \ref{fig12}.

\subsection{Ablation Experiments}\label{S6}
The above two subsections verify the feasibility of the trained policy networks of the Ackermann-steering leader and Omnidirectional followers. This section aims to give ablation experiments to demonstrate the advantage of the proposed HHy-PIDRL framework.  Given that the core component of HHy-PIDRL framework is the HM-DRL formation control law,  and the control law contains physical feedforward controller, PD feedback controller and adaptive RL residual controller. Similarly, triangular formation and vertical line formation are considered to conduct the  ablation experiments. The following four cases are considered:

$\textcircled{1}$: Adaptive RL controller. 

$\textcircled{2}$: Physical feedforward controller and adaptive RL residual controller. 

$\textcircled{3}$: PD feedback controller and adaptive RL residual controller.

$\textcircled{4}$: Physical feedforward controller, PD feedback controller, and adaptive RL residual controller.

\begin{table}[!htp]
\caption{statistical results of triangular formation} 
 \centering  
 \scalebox{0.8}{ 
\begin{tabular}{ccccc}
   \toprule
Test Results            & $\textcircled{1}$  &  $\textcircled{2}$ & $\textcircled{3}$ & $\textcircled{4}$ \\
   \midrule
Average Position Error  & $1.399\mathrm{m}$  & $0.312\mathrm{m}$    &   $0.600\mathrm{m}$               &                  $\mathbf{0.191m}$ \\
Average Success Rate    &   $90.00\% $        &                   $100\%$ &     $95\%$   &        $\mathbf{100\%}$ \\
Average Spacing Error   &    $0.626\mathrm{m}$      &   $0.160\mathrm{m}$    &      $0.145\mathrm{m} $            &                  $\mathbf{0.203m}$  \\
\bottomrule
\end{tabular}
}
\label{table7}
\end{table} 

TABLE \ref{table7} presents the average position error, average spacing error, and average success rate for followers to track the leader in a triangular formation pattern under four cases. From TABLE \ref{table7}, the metrics of the case that only  adaptive RL controller adopted  is obviously worse than the other three cases. Although the success rate under  case $\textcircled{2}$ and case $\textcircled{4}$  are $100\%$, the average position error and average spacing error under case $\textcircled{4}$  are obviously lower than them under the case  $\textcircled{2}$. In addition, the success rate under case $\textcircled{2}$ is higher than it under the case $\textcircled{1}$. From the above analysis, we conclude that the HM-DRL formation control law proposed in this paper is better than the other three situations in terms of success rate. 
 
\section{Conclusions}\label{S7}
To address the limitations of traditional control methods in handling model uncertainties and external disturbances, as well as the low sample efficiency and poor convergence of purely end-to-end RL approaches, this paper proposes a HHy-PIDRL framework. This framework aims to achieve high-precision and highly responsive formation control for HMRSs. The main conclusions are as follows: 1) Hierarchical Framework: We designed a two-layer architecture where the upper layer utilizes the SAC-based RL algorithm for the autonomous navigation of the Ackermann-steering leader. The lower layer introduces a novel HM-DRL formation control policy by integrating a high-fidelity physical feedforward controller, a classical PD controller, and an adaptive DRL residual controller. 2) Optimized Training: A hierarchical reward function was designed to effectively train the Omnidirectional followers. 3) Performance Validation: Experimental results demonstrate that both the upper-level autonomous navigation policy and the lower-level HM-DRL formation control policy achieved a $100\%$ success rate. Furthermore, ablation studies verified the validity and credibility of the proposed method. In summary, the HHy-PIDRL framework successfully synergizes the stability of physics-based control with the adaptability of deep reinforcement learning, providing a  solution for the complex formation control of HMRSs.

\end{document}